%% file: sn-article.tex
\documentclass[pdflatex,sn-mathphys-num]{sn-jnl}
\usepackage{titlesec}
\usepackage{adjustbox}
\usepackage{tabularx}
\newcolumntype{Y}{>{\centering\arraybackslash}X} 
\newcolumntype{L}{>{\raggedright\arraybackslash}X} 
\newcolumntype{C}{>{\centering\arraybackslash}X} 
\usepackage{graphicx}%

\usepackage{booktabs}%
\usepackage{multirow}%
\usepackage{amsmath,amssymb,amsfonts}%
\usepackage{amsthm}%
\usepackage{mathrsfs}%
\usepackage[title]{appendix}%
\usepackage{xcolor}%
\usepackage{textcomp}%
\usepackage{manyfoot}%
\usepackage{algorithm}%
\usepackage{algorithmicx}%
\usepackage{algpseudocode}%
\usepackage{listings}%
\usepackage[T1]{fontenc}
\usepackage{tcolorbox}
\usepackage{colortbl}
\usepackage{fontawesome5}
\usepackage{lipsum}
\usepackage{tikz}
\usetikzlibrary{shadows}
\usepackage{lmodern}

\usepackage{enumitem}

\usepackage{adjustbox}

\usepackage{makecell}
\usepackage{subcaption}

\usepackage{longtable}

\usepackage{times}

\definecolor{mySkyBlue}{RGB}{135, 206, 235}
\definecolor{boxcolor}{RGB}{70,130,180} 
\definecolor{textcolor}{RGB}{25,25,112} 
\definecolor{BerkeleyBlue}{HTML}{C41230}

\newtcolorbox{myinfobox}[2][]{
  colback=white,
  colframe=boxcolor,
  coltitle=white,
  fonttitle=\bfseries\sffamily,
  fontupper=\sffamily,
  top=3mm,
  bottom=3mm,
  left=3mm,
  right=3mm,
  boxrule=0.8pt,
  arc=7pt,
  title=#2,
  overlay={
    \fill[boxcolor!30] (frame.south west) -- (frame.north west) -- ([xshift=1cm]frame.north west) -- ([xshift=1cm]frame.south west) -- cycle;
  },
  #1
}

\newcommand{\modelname}{\textsf{PROTEUS\ }}

\newtcolorbox[use counter=examplecounter]{collapsibleexample}[1]{
  enhanced jigsaw,
  colback=white,
  colframe=gray!50,
  coltitle=black,
  fonttitle=\bfseries,
  title={Example \thetcbcounter: #1},
  fontupper=\normalsize,
  lower separated=false,
  before upper={\parindent15pt},
  collapse
}

\newcommand{\titlefont}{\color{black}\bfseries\fontsize{19}{20}\selectfont}

\titleformat{\section}
  {\normalfont\fontsize{16pt}{19.2pt}\bfseries}{\thesection}{1em}{}

\titleformat{\subsection}
  {\normalfont\fontsize{14pt}{16.8pt}\bfseries}{\thesubsection}{1em}{}

\usepackage{fancyhdr}
\usepackage{lipsum}

\begin{document}

\setlength{\parskip}{0.3em}

\renewcommand{\maketitle}{
\thispagestyle{plain}  
  \vspace*{0.5cm}  
  \hrule height 1pt  
  \vspace{2.4em}  
  \begin{flushleft}
    {\Large\bfseries\titlefont{Automating Exploratory Proteomics Research via Language Models}\par}
    \vspace{0.4cm}    
    
    {\normalsize
    \noindent
    \begin{tabular}{@{}l@{}}
      \textbf{Ning Ding}$^{1,2,*}$,
      \textbf{Shang Qu}$^{1,*}$,
      \textbf{Linhai Xie}$^{4,5,*}$,
      \textbf{Yifei Li}$^{1, 4}$,
      \textbf{Zaoqu Liu}$^{4,6}$,
      \textbf{Kaiyan Zhang}$^{1,3}$,
      \textbf{Yibai Xiong}$^{1,3}$, \\
      \textbf{Yuxin Zuo}$^{1}$,
     \textbf{ Zhangren Chen}$^{3}$,
      \textbf{Ermo Hua}$^{1,3}$,
    \textbf{Xingtai Lv}$^{1,3}$,
      \textbf{Youbang Sun}$^{1}$,
      \textbf{Yang Li}$^{4}$,
     \textbf{ Dong Li}$^{4}$, \\
      \textbf{Fuchu He}$^{4,5\dagger}$,
     \textbf{ Bowen Zhou}$^{1,2 \dagger}$,
    \end{tabular}
    \par}
    \vspace{0.5em}

    {\small
    \noindent
    \begin{tabular}{@{}l@{}}
      $^1$Tsinghua University, 
      $^2$Shanghai Artificial Intelligence Laboratory, 
      $^3$Frontis AI\\
      $^4$National Center for Protein Sciences (Beijing), State Key Laboratory of Medical Proteomics, Beijing Proteome Research Center \\
      $^5$International Academy of Phronesis Medicine (Guangdong) \\
      $^6$State Key Laboratory of Medical Proteomics, Beijing Proteome Research Center
    \end{tabular}
    \par}
    \vspace{0.5em}
    
    {\noindent \small * These authors contributed equally to this work.\par}
    {\noindent \small $\dagger$ Corresponding authors: \par}
  \end{flushleft}
}

\renewenvironment{abstract}
 {\par\noindent\textbf{Abstract.}\space}
 {\par\vspace{2em}\hrule height 1pt\vspace{1em}}  
 
\maketitle
\thispagestyle{empty}

\vspace{0.5cm}

\begin{abstract}

\noindent 
With the development of artificial intelligence, its contribution to science is evolving from \textit{simulating a complex problem} to \textit{automating entire research processes and producing novel discoveries}. Achieving this advancement requires both specialized general models grounded in real-world scientific data and iterative, exploratory frameworks that mirror human scientific methodologies. 
In this paper, we present \textsf{PROTEUS}, a fully automated system for scientific discovery from raw proteomics data. 
\modelname uses large language models (LLMs) to perform hierarchical planning, execute specialized bioinformatics tools, and iteratively refine analysis workflows to generate high-quality scientific hypotheses. 
The system takes proteomics datasets as input and produces a comprehensive set of research objectives, analysis results, and novel biological hypotheses without human intervention. We evaluated \modelname on 12 proteomics datasets collected from various biological samples (e.g. immune cells, tumors) and different sample types (single-cell and bulk), generating 191 scientific hypotheses. These were assessed using both automatic LLM-based scoring on 5 metrics and detailed reviews from human experts. Results demonstrate that \modelname consistently produces reliable, logically coherent results that align well with existing literature while also proposing novel, evaluable hypotheses. The system's flexible architecture facilitates seamless integration of diverse analysis tools and adaptation to different proteomics data types. By automating complex proteomics analysis workflows and hypothesis generation, \modelname has the potential to considerably accelerate the pace of scientific discovery in proteomics research, enabling researchers to efficiently explore large-scale datasets and uncover biological insights.

\end{abstract}

\section{Introduction}\label{sec1}

Proteomics research~\cite{aslam2016proteomics}, which focuses on the large-scale analysis of protein expression, functions, and interactions, is a crucial avenue for understanding biological processes and their underlying mechanisms. Modern technologies~\cite{bennett_single-cell_2023} have facilitated high-throughput proteomics sequencing and large-scale data collection. The resulting datasets hold copious information on proteins, cells, pathways, as well as their complicated relationships and interactions. 
When combined with scientific analysis methods and domain knowledge, they have the potential to reveal valuable biological insights, including novel biomarkers~\cite{schmidt_plasma_2024}, disease mechanisms~\cite{sasvari_neutrophil-specific_2024}, and therapeutic targets~\cite{turkki_tensin-2_2024}. On the other hand, the sheer volume and complexity of proteomics data also pose challenges for conventional research techniques and paradigms.
Current proteomics research relies heavily on human experts to design and perform data analysis using professional methods and tools, making decisions ranging from specific data manipulation to general research directions. 
This process brings forward two main issues. First, analysis can be extremely time-consuming, especially when it involves trial-and-error over large sets of possible proteins or sample groups. Second, the researcher's personal knowledge and habits may bias experimental design, potentially impeding comprehensive analysis and limiting its overall scope.

We propose that large language models (LLMs)~\cite{achiam2023gpt, dubey2024llama, han2021pre, bommasani2021opportunities}, the cornerstone of generative artificial intelligence, can enable unprecedented extents of automation in proteomics research. 
State-of-the-art LLMs possess powerful instruction-following abilities and extensive general knowledge, which have expanded their use cases from simple language tasks to a myriad of professional domains~\cite{saab2024capabilities,zhang2024ultramedical}. 
They have also demonstrated impressive competence and flexibility in realms such as planning complex tasks and calling diverse tools~\cite{qin_tool_2024, m_bran_augmenting_2024}. 
Additionally, for knowledge-intensive or time-sensitive scenarios, augmenting LLMs with information retrieved from external sources effectively reduces hallucinations and improves accuracy~\cite{xiong_benchmarking_2024, jin_genegpt_2024}.
The primary advantage of LLMs over previous machine learning approaches for scientific tasks is their versatility: instead of being confined to a narrowly defined problem, LLMs can simulate a broad range of tasks integral to scientific discovery workflows. In other words, through representing all decision-making steps, intermediate results, and reasoning processes as sequences of tokens, we enable LLMs to generalize across these disparate steps within the research process.
In the context of proteomics research, the generalization capabilities of LLMs, combined with specific data, information, and tools, can enable automated systems to advance from surface-level data analysis to in-depth scientific hypothesis proposal.
This allows for more efficient, comprehensive, and insightful explorations of high-throughput proteomics data, and mitigates human experts' possible biases by uncovering hypotheses they might have overlooked.  
With this objective, we envision an end-to-end proteomics analysis and knowledge discovery system, where the input is raw data and the output is a set of scientific hypotheses derived from that data.

In this paper, we develop a fully automated LLM-based \textbf{PROT}eomics \textbf{E}xploration and \textbf{U}nderstanding \textbf{S}ystem (\textbf{PROTEUS}) for proteomics analysis and scientific knowledge discovery.
\modelname first receives raw omics data and basic dataset information, based on which it plans data-dependent analysis procedures across three hierarchies: research objectives, analysis workflows, and analysis tools. Guided by these plans, the system automatically executes a sequence of analysis steps using bioinformatics tools, then interprets the results. Furthermore, \modelname performs iterative refinement after completing each workflow or objective, updating subsequent plans based on the latest results. We demonstrate the capabilities of \modelname through both automatic and human evaluation. \modelname automatically analyzed 12 diverse proteomics datasets and produced a total of 191 high-level scientific hypotheses. We first used LLMs to score each hypothesis based on 5 metrics, with access to supplementary information such as the original research paper and related articles. To validate the automatic scoring results, we randomly selected a subset of results and obtained evaluation scores from human experts using the same metrics and instructions. Experts also provided detailed, open-ended feedback on the quality, novelty, and biological implications of the hypotheses. We show that \modelname can flexibly explore different types of proteomics data, consistently producing high-quality and novel scientific hypotheses.

Previous research on using LLMs to enhance omics research has largely been confined to isolated steps within the complex research process. Common tasks include batch effect correction~\cite{xiao_cellagent_2024}, cell type annotation~\cite{xiao_cellagent_2024}, and differential gene selection~\cite{zhou_ai_2023}. While these methods provide valuable assistance to bioinformatics researchers, they lack the flexibility and comprehensiveness required for automating entire omics research procedures. In contrast, we enable \modelname to freely employ diverse methods and directly derive scientific hypotheses from raw data, bringing the system's performance closer to the iterative, exploratory research process of human experts. Most existing methods that similarly encompass the full bioinformatics research pipeline still rely heavily on human intervention, either requiring a pre-determined procedure to link single steps~\cite{liu_data-intelligence-intensive_2024}, or relying on frequent user inputs to guide the analysis~\cite{xin_bioinformatics_nodate}~\cite{lu_scchat_2024}. DREAM~\cite{deng_dream_nodate}, while eliminating human inputs, evaluates the system's outputs solely by judging whether the initial research question was resolved, lacking verification of the reliability and depth of the results. We address this gap by jointly employing human evaluation and a well-rounded suite of automatic evaluation methods and metrics that align with the open-ended nature of \modelname's outputs. Therefore, our work represents a pioneering effort to incorporate both proteomics research and result evaluation in a fully automated, end-to-end manner, advancing AI-assisted efficient bioinformatics research.

\section{Results}\label{sec2}

\subsection{\modelname System}

Towards the goal of automated proteomics research from raw data, we develop \modelname, a system that combines the general abilities of language models with the accuracy of domain-specialized analysis tools and knowledge sources. The system's input is a proteomics dataset consisting of protein expression data and cell or sample metadata. A large language model orchestrates the analysis process and arrives at a list of specific, data-grounded scientific hypotheses.

\begin{figure}
    \centering
    \includegraphics[width=0.95\linewidth]{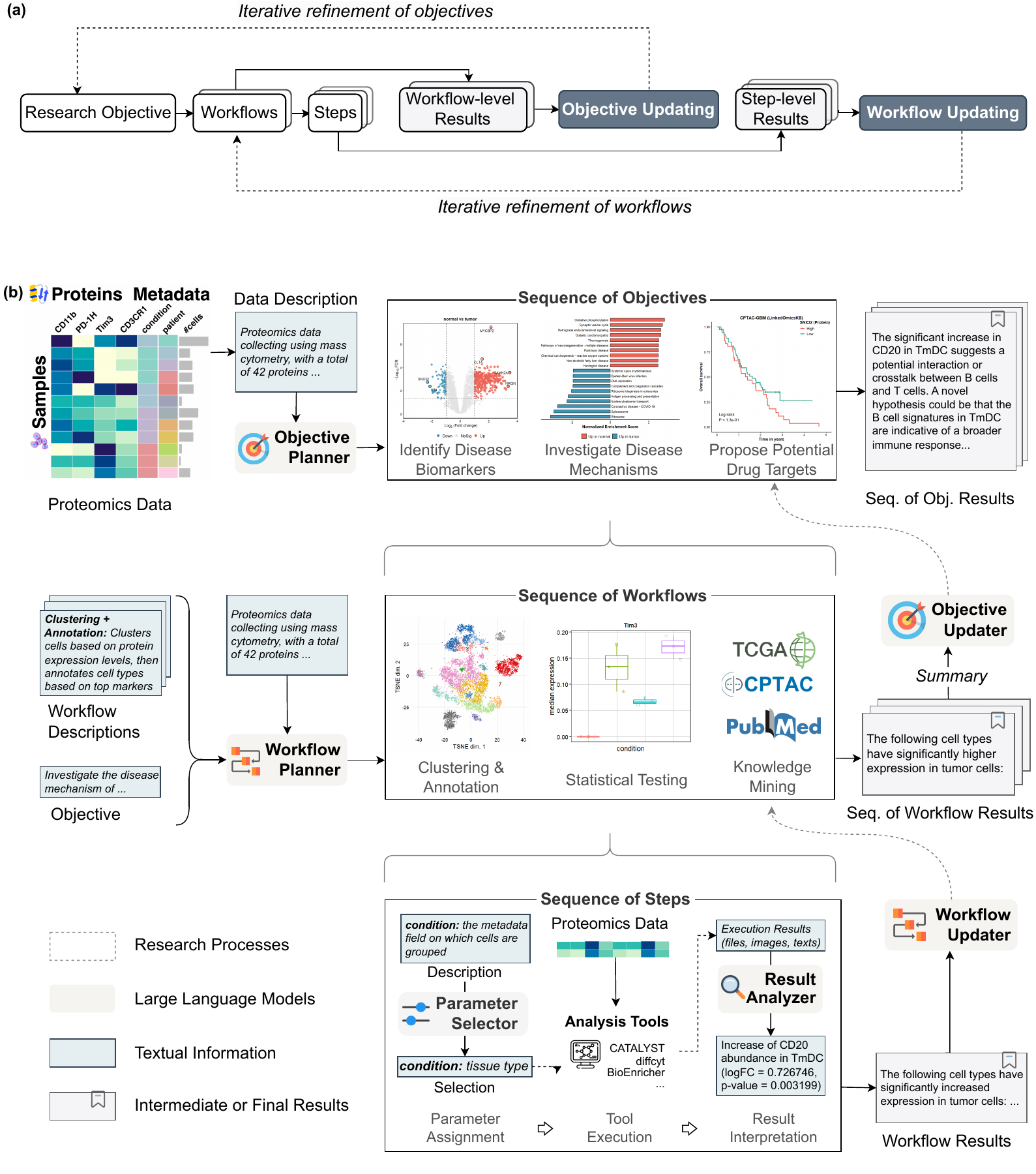}
    \vspace{0.3cm}
    \caption{(a) The framework of the iterative refinement of \modelname. (b) A detailed illustration of the working process of \modelname.  
    First, the Data Description module generates a precise description of the proteomics dataset, based on which the Objective Planner proposes a series of research objectives. Based on each objective, the Workflow Planner then plans a sequence of analytical workflows, such as analyzing expression differences or labeling cell types. These planned workflows are executed using specialized bioinformatics tools and follow a fixed sequence of steps. The Workflow Updater and Objective Updater analyze the system's latest results, based on which they refine the subsequent workflows and objectives. \modelname produces numerous scientific hypotheses for each research objective, and continues its analysis until it reaches a pre-determined maximum number of objectives. These framework designs facilitate a robust, end-to-end proteomics research pipeline.}
    \vspace{-0.3cm}
    \label{fig:framework}
\end{figure}

\subsubsection{Large Language Models}

Currently, the gap between proprietary and open-source language models is narrowing, with both showing capabilities for executing complex planning and reasoning tasks. Given the confidential and privacy-sensitive nature of proteomics data, as well as the need for iterative model improvements, we train a general-purpose model with a focus on the biomedical field. Specifically, we train models on biomedical instruction datasets to enhance their capabilities for analysis, planning, and knowledge in biomedicine, thereby improving their performance within our proteomics scientific discovery system. We predominantly adopt the state-of-the-art open-source LLM, Llama 3.1~\cite{dubey2024llama}, as our backbone architecture. With 70 billion parameters, it outperforms previous open-source models~\cite{achiam2023gpt} across a range of tasks. For further domain specialization, we fine-tuned Llama 3.1 70B on the UltraMedical dataset~\cite{zhang2024ultramedical}, which contains diverse and high-quality biomedical instructions, including a wealth of open-ended questions on biomedical research and literature. The resulting models demonstrate superior performance on downstream tasks compared to other open-source models.

\subsubsection{System Framework}

\modelname arrives at a comprehensive and meaningful set of hypotheses through navigating possible objectives, executing statistical analysis, and iteratively improving its analysis plans. Considering the complexity and diversity of proteomics research, we devise a hierarchical planning framework consisting of three levels, ranging from general to specific: research objectives, analysis workflows, and analysis steps. This design increases flexibility and robustness in the LLM planning process, which forms the backbone of \modelname. We also incorporate self-refinement and hypothesis proposal steps to further improve result quality. We detail the design of each module below.

\input{table-workflow-info}

\noindent \textbf{Research Objectives: Guiding the Trajectory of Proteomics Exploration.} Proteomics research encompasses diverse objectives, such as establishing protein interactions, elucidating disease mechanisms, and identifying disease biomarkers. These high-level objectives determine the direction of data analysis and hypothesis proposal.
In \modelname, we take advantage of the planning and reasoning capabilities of LLMs to dynamically generate and refine these research objectives.
The LLM first generates a description of the input data, covering both protein expression data and relevant metadata, providing a comprehensive overview of the dataset. It includes information such as the number of proteins and cells sequenced, the conditions of the cell samples, and other important metadata paired with their possible values.
Given the data description, the LLM proposes several potential research objectives to be considered sequentially and is encouraged to tailor them to data characteristics. 
For instance, it may highlight important cell markers or cell types based on general knowledge of the disease conditions mentioned in the description.

\noindent \textbf{Analysis Workflows: Streamlining Complex Bioinformatics Processes.} Due to the rigorous dependencies and high specificity of many bioinformatics data analysis methods, we organize a large number of analysis tools into a set of analysis workflows, each consisting of one or more tools to be executed in sequence, as well as additional steps necessary for the analysis process. 
One example is the cell type annotation workflow, which calls the cell clustering tool, then the cluster annotation tool. 
For a certain research objective, we prompt the LLM with the objective, the data description, and a list of descriptions of all available data analysis workflows, then instruct it to plan a series of workflows. This design greatly reduces the probability that the system encounters errors caused by tool or data dependencies. It also reduces the difficulty and complexity of planning by reducing the total number of options given to the LLM.

\noindent \textbf{Analysis Steps: Enabling Professional, Data-Grounded Proteomics Analysis.} Each analysis workflow includes both bioinformatics tools and additional analysis steps conducted by the LLM, which serves as an orchestrator for tool-related functionalities. We focus on two critical tasks: determining optimal tool parameters and interpreting complex execution results.
To automatically set tool parameters, the LLM is provided with the research objective, the data description, and an explanation of each parameter. For interpreting results, \modelname supports various formats of tool outputs, including text, data files, and visualization plots, and analyzes notable results within the context of the research objective. For example, after performing differential abundance analysis, \modelname summarizes key insights and biological implications based on the result file, identifying and emphasizing cell types that exhibit statistically significant changes and are related to the objective. The LLM is capable of providing further in-depth analysis, for instance regarding the functions of the notable cell types and the biological implications of their changes.

\noindent \textbf{Hierarchical Iterative Refinement.} Proteomics research is an iterative process in which results from preliminary analysis stages can be conducive to deeper and more detailed exploration. Therefore, we enable \modelname to refine its plans after each execution stage. Following each workflow execution, the LLM refers to the newly obtained workflow results to update the original plan in preparation for subsequent workflow execution. It performs a similar step after analyzing each objective, using the latest results to refine future research objectives. These additional steps assist \modelname in both handling errors and deepening scientific inquiry. For instance, if initial analysis reveals that a particular cell type exhibits significant changes in abundance, the updated research objectives may propose identifying more fine-grained subtypes to clarify the observed trends.

\noindent \textbf{Hypothesis Proposal.} Finally, \modelname proposes hypotheses from the completed analysis through integrating information on research objectives, executed workflows, and result interpretations. We instruct the LLM to emphasize the most significant and novel results among a large number of possible directions. The prompt specifies a fixed format for each hypothesis, consisting of an overview of proteins and cell conditions, a summary of statistical tests and corresponding numerical results, and a final scientific hypothesis. As a result, the summaries effectively link all proposed hypotheses to direct statistical observations derived from raw input data. This increases the interpretability of \modelname's outputs and enables effective evaluation.

\subsubsection{Features of \modelname}

\noindent \textbf{From Raw Data to Scientific Discoveries.} 
\modelname exemplifies a paradigm shift in bioinformatics research by achieving a fully automated pipeline that produces scientific discoveries from raw data. Unlike traditional methods that rely heavily on human intervention and manual data processing, \modelname leverages the capabilities of LLMs to autonomously navigate the complete research process. This end-to-end pipeline ensures consistency, reduces the potential for human error, and significantly reduces the time spent between data acquisition and hypothesis generation.

\noindent \textbf{Scalable Integration.} 
Due to its hierarchical planning framework, \modelname can seamlessly integrate diverse proteomics analysis tools and knowledge sources. The three-tiered structure, encompassing research objectives, analysis workflows, and individual tools, enables flexible adaptation to heterogeneous proteomics datasets and diverse research directions. Therefore, \modelname can conveniently incorporate new analysis methods and external data while maintaining a fixed system framework.
Furthermore, the LLM's role in parameter assignment and result interpretation allows \modelname to execute specialized bioinformatics tools while maintaining a unified interface. This approach of scalable integration positions \modelname to evolve alongside advancements in bioinformatics methods and technologies, ensuring its long-term effectiveness.

\noindent \textbf{Dynamic Feedback Loop.}
\modelname implements an iterative refinement process that mimics the recursive process of scientific inquiry commonly employed by human experts. After each workflow execution and objective analysis, the system reevaluates and refines its subsequent plans based on newly obtained results. This dynamic approach allows \modelname to adapt to unexpected findings and pursue promising avenues of research that may not have been initially apparent. By incorporating feedback loops at multiple levels of analysis, \modelname can conduct thorough and nuanced investigations, uncovering insights that might be overlooked under linear analysis approaches.

\subsection{Evaluation}

\subsubsection{Base Language Models}

We aim for the base language model of the system to be a specialized generalist, achieving enhanced specialization in the biomedical field without compromising its generalization abilities on common tasks.
We present the evaluation results of our custom Llama 3.1 models tailored to UltraMedical specifications. 
We primarily base our evaluation methodology on the protocols outlined in~\cite{zhang2024ultramedical} and assess the models across widely recognized medical and general benchmarks.
For medical benchmarking, we selected MultiMedQA, which has been extensively employed in MedPaLM-related studies~\cite{singhal2023large,singhal2023towards}. This benchmark comprises MedQA~\cite{jin2021medqa}, PubMedQA~\cite{jin2019pubmedqa}, MedMCQA~\cite{pal2022medmcqa}, and biomedical categories within MMLU~\cite{hendrycks2020mmlu}.
We select these tasks to assess the LLMs' application of biomedical knowledge. Additionally, for general instruction following and knowledge integration, we primarily evaluate the models across the comprehensive set of MMLU, GPQA~\cite{rein2023gpqa}, and Alpaca Eval 2~\cite{dubois2024alpaca} benchmarks.

The overall results are listed in Table~\ref{tab:general_domain_results}, and the detailed performance in medical domain is reported in Table~\ref{tab:med}.
Our model demonstrates impressive performance across both biomedical and general domain benchmarks. On the MultiMedQA benchmark, our model achieves an average accuracy of 86.30\%, surpassing other biomedical-focused models such as Med42-70B (70.74\%), OpenBioLM-70B (86.06\%), and Med-PaLM 2 (ER) (85.46\%). 
Notably, it also outperforms general domain models including GPT-3.5-Turbo (67.80\%) and comes close to GPT-4-Turbo (87.00\%). 
On the Alpaca Eval 2 benchmark, our model shows strong performance with a win rate (WR) of 46.09\% and a Likert score (LC) of 43.45\%, considerably outperforming other biomedical models and many general domain models. The MMLU benchmark presents similar results.
On the GPQA benchmark, our model demonstrates an accuracy of 45.76\%, competitive with top-performing general models such as GPT-4-Turbo (49.10\%) and Llama-3.1-70B-Instruct (46.70\%). 
Our model acts as the core orchestrator within \modelname, supporting its comprehensive planning and reasoning, leading to novel scientific hypotheses.

\input{table-llm-auto}

\input{table-llm-medqa}

\input{table-stats}

\subsubsection{Quantitative Evaluation}

We conducted experiments and quantitative evaluation on two types of proteomics data. First, we used the Single-cell Proteomic DataBase (SPDB)~\cite{wang2024spdb} to obtain 10 single-cell datasets which used cytometry by time-of-flight (CyTOF)~\cite{bandura_mass_2009} sequencing technology. 9 datasets were sequenced on various human tissues, and 1 dataset covered mouse brain tumor tissue. For all experiments on SPDB, \modelname's input was a \textsf{SingleCellExperiment}~\cite{amezquita_orchestrating_2020} object directly downloaded from SPDB and a textual data description constructed using information from the data object and the SPDB website. For every dataset, we set the maximum number of total research objectives to 3 and instructed \modelname to generate 5 hypotheses for each objective. On several objectives, \modelname produced less than 5 hypotheses due to a lack of notable results from the analysis. We collected a total of 147 hypotheses for CyTOF data. In addition, to demonstrate the flexibility of \modelname, we obtained two clinical proteomics datasets from previous publications on hepatocellular carcinoma (HCC)~\cite{jiang2019proteomics} and glioblastoma (GBM)~\cite{oh2020integrated}. These datasets contain bulk proteomics data sequenced using mass spectrometry (MS)~\cite{domon_mass_2006} and cover significantly larger numbers of proteins than the previously described SPDB datasets. The system's input for each dataset was 2 files containing the raw protein

\input{table-average-scores}

\noindent  expression data and clinical feature metadata respectively. \modelname produced a total of 44 hypotheses, 20 on the HCC dataset and 24 on the GBM dataset. All of the above results were produced fully automatically, without human intervention.

For the two clinical proteomics datasets, we adjusted the system in two main ways to address the differences in data characteristics. First, we allowed \modelname to directly call individual tools without introducing workflows. This is because bioinformatics analysis on clinical proteomics data is generally more flexible, with less reliance on fixed analysis pipelines. Second, since this data type requires distinct analysis methods, we replaced the set of analysis tools available to \modelname with bioinformatics tools commonly used for clinical proteomics data. Details on all analysis workflows and tools are provided in Table~\ref{tab:workflow_info}.
The statistics of the datasets and hypotheses produced by our system are listed in Table~\ref{tab:stats}.
In Table~\ref{tab:workflow-frequency}, we provide the frequency each workflow was called during one experimental run over 10 CyTOF datasets.

We used the state-of-the-art large language model GPT-4o to assist automatic quantitative evaluation. We evaluated each hypothesis separately according to 5 distinct metrics by prompting the LLM with the full hypothesis and optional refererence information. The prompts outlined detailed scoring criteria for the metrics and instructed GPT-4o to provide free-form analysis, followed by an integer score between 0 and 5. 
The 5 evaluation metrics are: Paper-Based Alignment, Literature-Based Alignment, Literature-Based Novelty, Logical Coherence, and Evaluability. These metrics are informative indicators of the plausibility, novelty, and potential for further exploration of a scientific hypothesis.

We next discuss the general evaluation results and provide detailed evaluation prompts in ~\ref{method-evaluation}. 
Table~\ref{tab:overview} shows the overall scores of the 191 hypotheses, and
Figure~\ref{fig:auto-scores} provides score distributions on all hypotheses. 
Table~\ref{table:data_scores} lists the different results on each dataset.

\renewcommand{\figurename}{Fig.}
\renewcommand{\thefigure}{2}

\begin{figure}
    \centering
    \includegraphics[width=\linewidth]{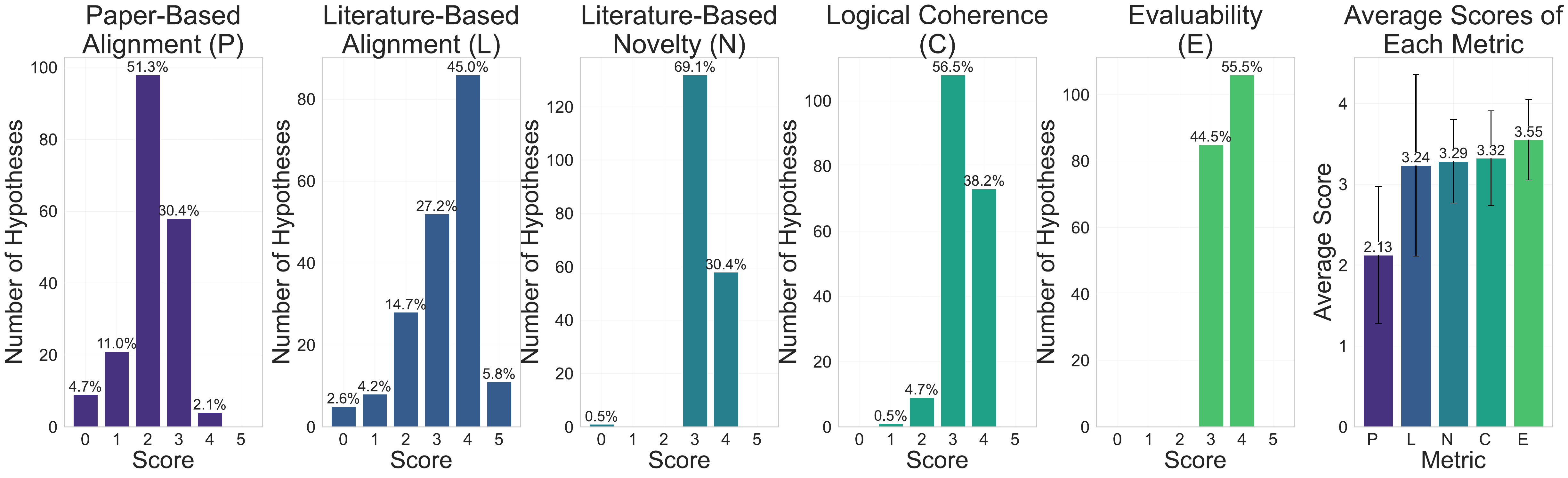}
    \caption{Average scores and score distributions calculated on all 191 hypotheses, over 5 metrics.}
    \vspace{-0.5cm}
    \label{fig:auto-scores}
\end{figure}

\input{table-dataset-scores}

\noindent \textbf{Evaluation based on corresponding published research.}
For each dataset, we accessed the original research paper that published and analyzed the data. We extracted the text and located the most relevant text chunk for each hypothesis, after which

\renewcommand{\figurename}{Fig.}
\renewcommand{\thefigure}{4}
\begin{wrapfigure}{r}{0.5\textwidth}
    \centering
    \includegraphics[width=1\linewidth]{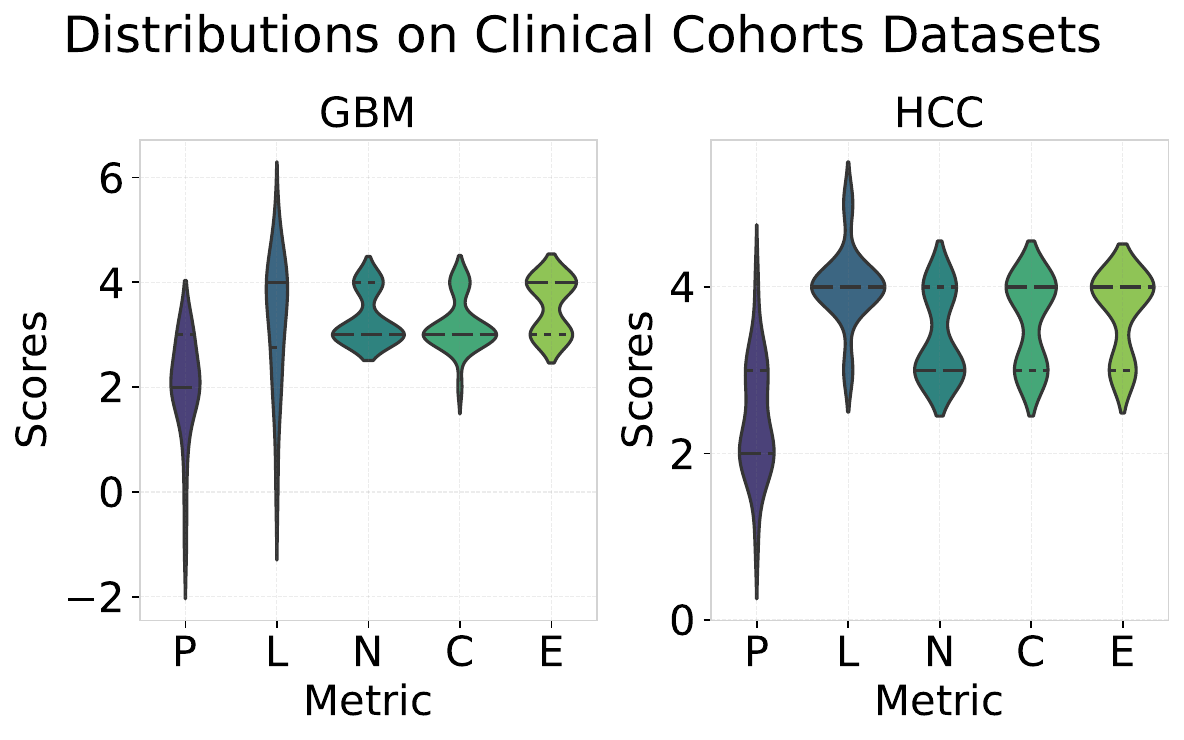}
    \caption{Score distributions on the 2 clinical cohort datasets (HCC and GBM), over 5 metrics.}
    \vspace{-0.8cm}
    \label{fig:distribution-clinical}
\end{wrapfigure}

\noindent GPT-4o evaluated the degree of overlap. We note that most proteomics datasets contain substantial amounts of information and potentially insightful trends, out of which only a subset is elaborated in the paper. \modelname is designed to conduct analyses in a comprehensive and unbiased manner, considering all possible research directions. Therefore, we expect that most responses will fall outside the scope of the original paper. A low score on this metric likely indicates discrepancies between the research focuses of \modelname and those of the original paper, rather than reflecting low quality of the results.

Scores on this metric ranged from 0 to 5, with an average of 2.13, meaning that most hypotheses proposed by \modelname were not covered in the original research papers.

\renewcommand{\figurename}{Fig.}
\renewcommand{\thefigure}{3}
\begin{figure}
    \centering
    \includegraphics[width=1\linewidth]{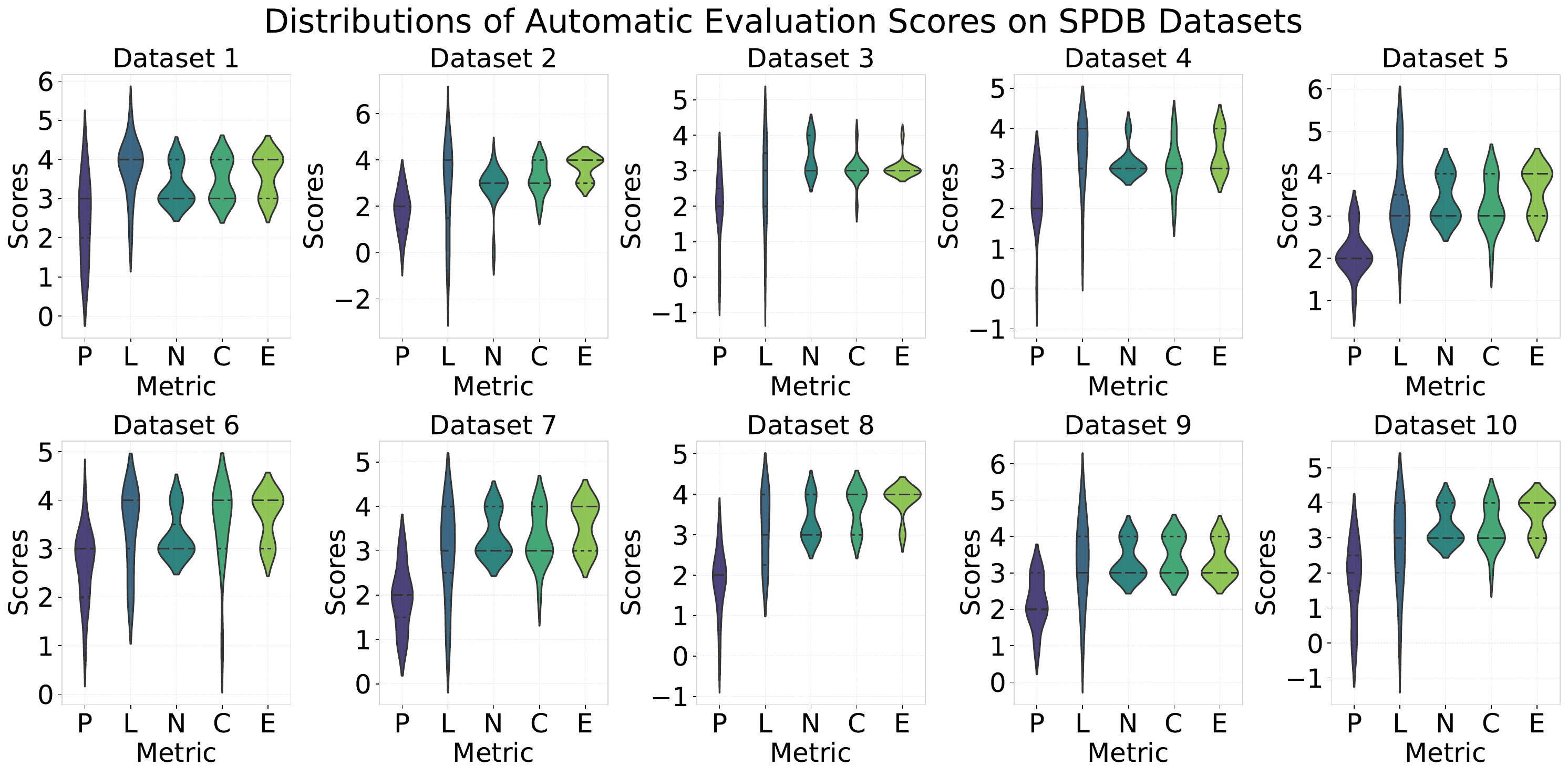}
    \caption{Score distributions on the 10 SPDB datasets, over 5 metrics. Specific dataset information corresponding to each index is provided in Table~\ref{tab:dataset-info}.}
    \label{fig:distribution-spdb}
\end{figure}

\noindent \textbf{Evaluation based on general literature.}
We next introduce two evaluation metrics that compare \modelname's outputs against general biological literature. For each hypothesis, we automatically searched 10-20 of the most relevant articles on PubMed and extracted their PMIDs, titles, and abstracts. The search term was a concise query generated by GPT-4o based on the full output, comprising the most important biological entities present in the output. Using this information, GPT-4o individually scored the generated hypotheses according to two considerations: 1) Literature-Based Alignment, which assesses the degree of alignment between the generated results and existing research; and 2) Literature-Based Novelty, which evaluates their originality within the context of current biological literature.

\modelname scored an average of 3.24 on Literature-Based Alignment and 3.29 on Literature-Based Novelty. Hypotheses consistently scored 3 or 4 on Novelty. On the other hand, results for Literature-Based Alignment showed the largest deviation among all 5 metrics, with scores ranging from 0 to 5. Nonetheless, more than 78\% of hypotheses scored at least 3 points, indicating that most results were partially concordant with existing research.

\noindent \textbf{Direct evaluation.}
Finally, we incorporate two metrics that can be evaluated solely based on the generated hypotheses, without additional information: 1) Logical Coherence, which assesses whether the output is logically plausible considering general biological principles; and 2) Evaluability, which determines whether the hypothesis can be further evaluated and verified through existing statistical methods or experimental procedures in the field. The LLM relies on its general domain knowledge and reasoning skills to judge the output.

\modelname's results reliably reached satisfactory scores, despite the evaluation system being relatively strict. Average scores for Logical Coherence and Evaluability were 3.32 and 3.55 respectively. Notably, all hypotheses had Evaluability scores of 3 or 4, and only 5.24\% of hypotheses had Logical Coherence scores of 2. In other words, an overwhelming majority of hypotheses were biologically plausible and at least partially evaluable.

\renewcommand{\figurename}{Fig.}
\renewcommand{\thefigure}{6}
\begin{wrapfigure}{r}{0.5\textwidth}
    \centering
    \vspace{-0.5cm}
    \includegraphics[width=0.8\linewidth]{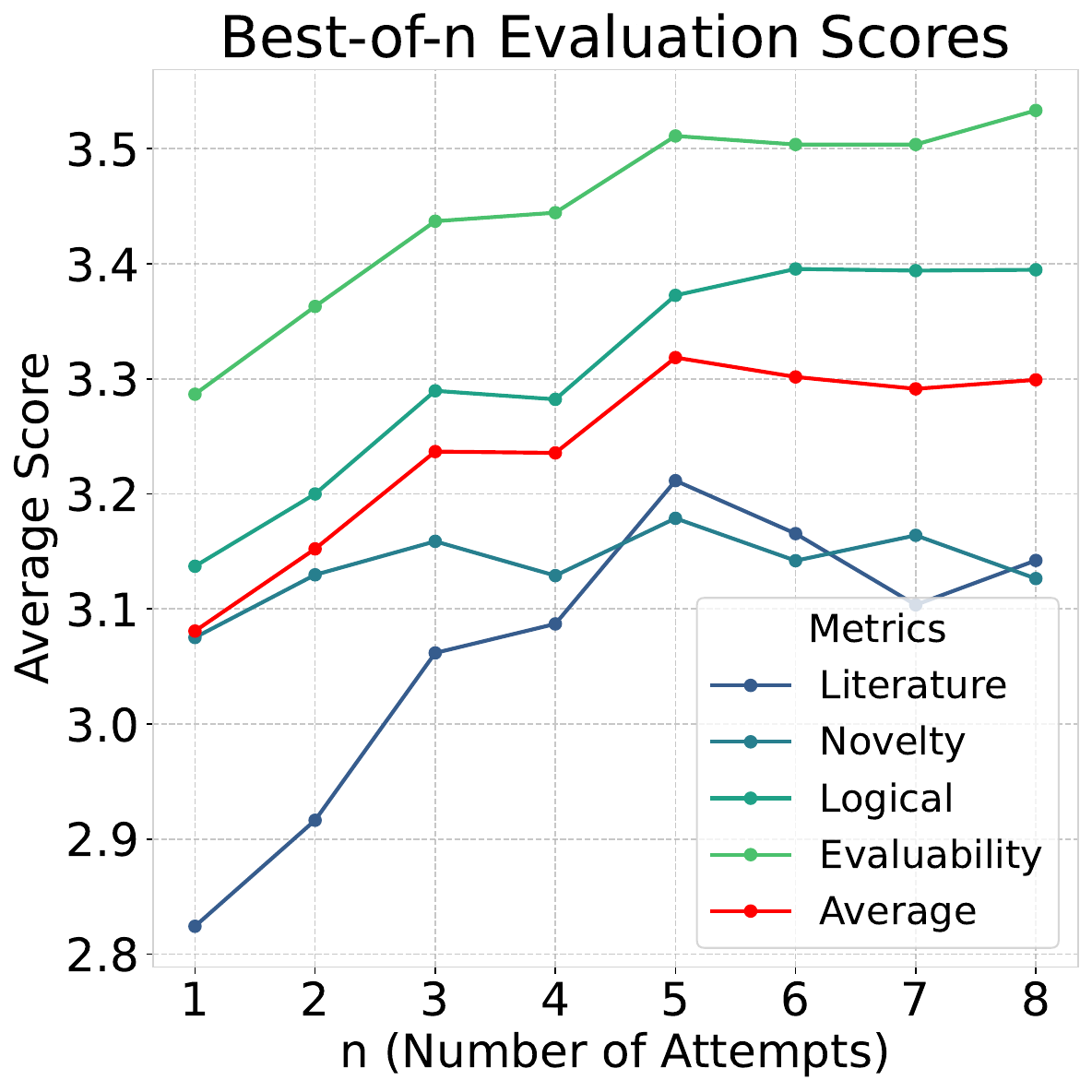}
    \caption{Results obtained by performing multiple experimental runs and automatically selecting the best hypothesis.}
    \vspace{-0.5cm}
    \label{fig:best-n}
\end{wrapfigure}

\noindent \textbf{Comparison of base language models.}
To evaluate the impact of the choice of LLMs in our framework, we conducted experiments on the 10 CyTOF datasets using GPT-4o as the base LLM, keeping the system framework and available workflows unchanged. We performed automatic evaluation using the same prompts and present the results in Figure~\ref{fig:base-models}. Results were mixed across the 5 different metrics, with GPT-4o achieving a slightly higher average score. Nonetheless, as the backbone model of \modelname, our model demonstrated competitive performance compared to the state-of-the-art proprietary model.

\renewcommand{\figurename}{Fig.}
\renewcommand{\thefigure}{5}
\begin{figure}
    \centering
    \includegraphics[width=\linewidth]{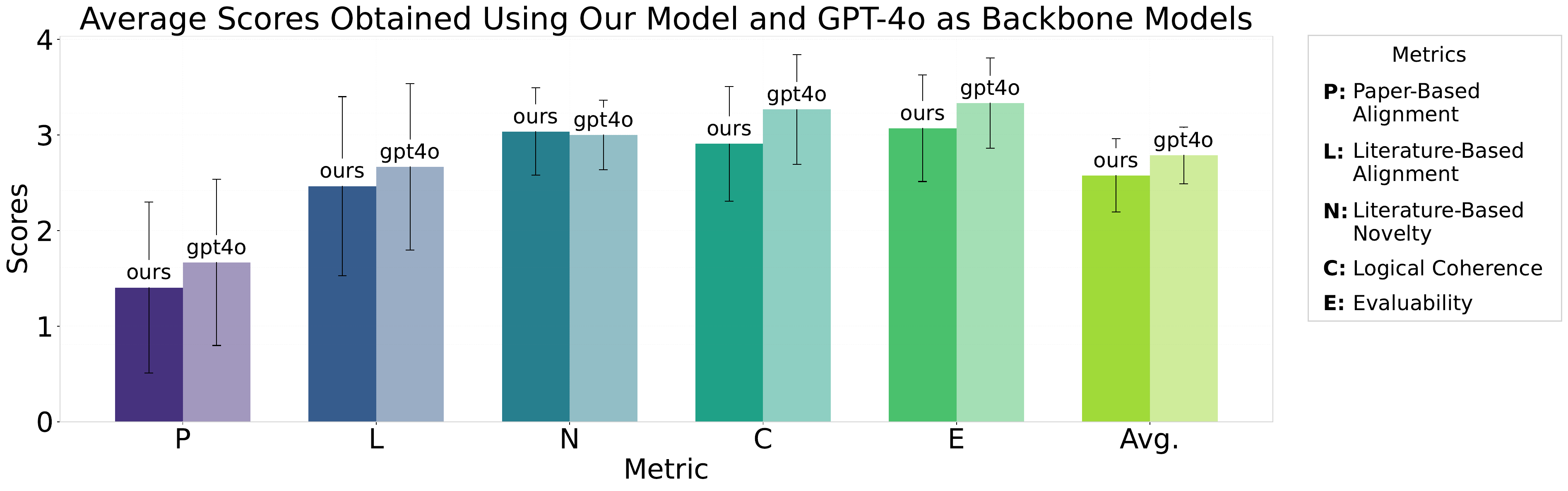}
    \caption{Comparison of results obtained using our model and GPT-4o as the backbone of \modelname.}
    \label{fig:base-models}
\end{figure}

\noindent \textbf{Performing multiple experimental runs.}
Due to the randomness in sampling outputs from an LLM, \modelname produces slightly different results for the same dataset with each run. Therefore, we investigated whether obtaining multiple sets of outputs and then automatically selecting the best hypotheses could improve the overall scores.

We present the results in Figure~\ref{fig:best-n}. Since degree of alignment with the original paper is a poor indicator of overall hypothesis quality, we calculated the average score across the remaining four metrics. All four metrics exhibited a general upward trend, with the average score improving by more than 0.2 starting from 5 iterations. These findings, combined with the efficiency of \modelname, demonstrate a promising approach for further enhancing result quality.

\subsubsection{Verifying Evaluation Quality.}

\noindent \textbf{Comparison with human scoring.} We randomly selected two datasets (Datasets 3 and 4) from SPDB, corresponding to 30 hypotheses. Human experts in proteomics research scored these hypotheses over the 5 metrics, following the same instructions provided to the LLM evaluator. These results, shown in Figure~\ref{fig:agreement}, demonstrate the rigor and validity of automatic scoring.
For all metrics except Novelty, the average scores given by human evaluators were higher than those from LLM automatic evaluation, indicating that automatic evaluation was generally more sensitive to minor errors or discrepancies. In addition, automatic and human scoring showed reasonable levels of agreement across all metrics.

\renewcommand{\figurename}{Fig.}
\renewcommand{\thefigure}{7}
\begin{figure}
    \centering
    \includegraphics[width=\linewidth]{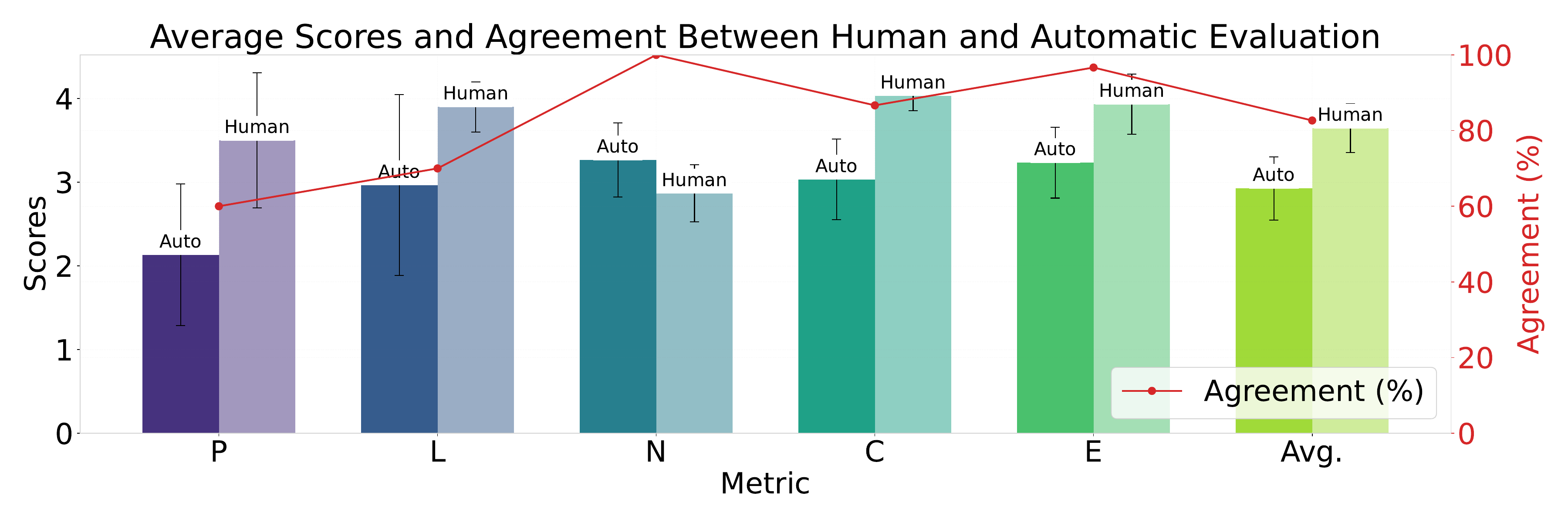}
    \caption{Comparison of automatic and human scoring results. The bar chart shows the average scores on each metric and the line chart denotes the agreement within a 1 point difference.}
    \label{fig:agreement}
\end{figure}

\renewcommand{\figurename}{Fig.}
\renewcommand{\thefigure}{8}
\begin{figure}
    \centering
    \includegraphics[width=\linewidth]{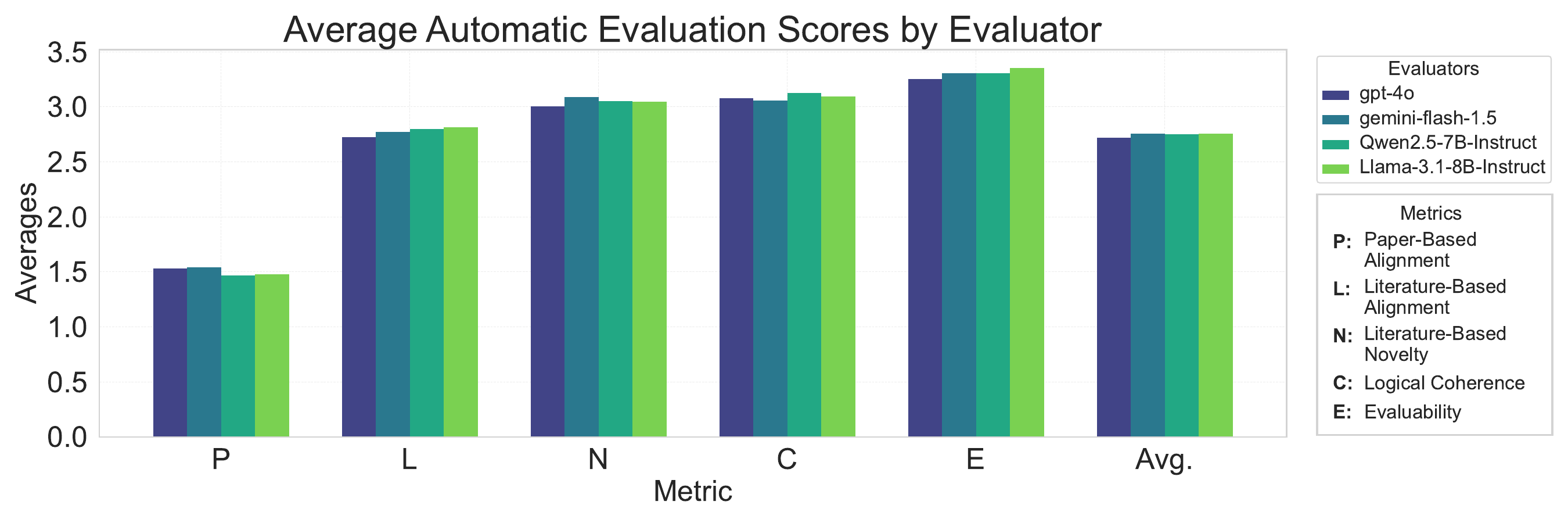}
    \caption{Comparison of automatic scoring results using four different LLM evaluators.}
    \label{fig:evaluators}
\end{figure}

\noindent \textbf{Comparison of different evaluator LLMs.}
In all previous experiments, we used GPT-4o as the automatic evaluator. We subsequently reran the evaluation procedure on all SPDB results using three other language models as evaluators: Gemini-flash-1.5, Qwen2.5-7B-Instruct, and Llama-3.1-8B-Instruct. We present the average scores on the 5 metrics and the overall average scores in Figure~\ref{fig:evaluators}. The resulting scores were close on all metrics, particularly for the overall average score, confirming the robustness of our automatic evaluation procedure.

\subsubsection{Case Studies}
\label{case-studies}

We highlight \modelname's ability to propose insightful and novel hypotheses through in-depth evaluation by human experts. We selected subsets of hypotheses from both types of proteomics data. Human experts reviewed the stated statistical trends and resulting hypotheses with reference to the original research paper and dataset. In the following section, we expand on several notable hypotheses to demonstrate \modelname's advantages as well as limitations. 

\noindent \textbf{\modelname identifies trends on rarely-studied biological topics and proposes novel hypotheses.} In our experiments, \modelname often focused on proteins or cell types that were seldom studied in the considered context, enabling novel and valuable results. On Dataset 2, the system observed significant changes ($\log_2 \text{FC} = 4.812894$, $\text{adjusted p-value} = 0.007$) in mesothelioma cells between different stages of High-Grade Serous Ovarian Cancer (HGSC) and suggested that "the increase in mesothelioma cells from III A to IV stages could indicate a potential therapeutic target."

Human experts provided positive feedback regarding the system's focus on the relationship between mesothelioma cells and HGSC, and judged the hypothesis as plausible, novel, and interesting. They noted that mesothelioma cells have seldom been studied in HGSC research. Further research via PubMed revealed less than 10 previous works that directly discussed this relationship, exploring the roles of the methothelial barrier~\cite{steitz_trail-dependent_2023}, methothelial cells~\cite{ghoneum_exploring_2021}, and mesothelial genes~\cite{ojasalu_upregulation_2020, coelho_regulation_2020}, in HGSC progression and clinical outcomes. None of the existing works directly overlapped with \modelname's proposed hypothesis. The biological plausibility of the hypothesis is supported by mesothelioma cells' known ability to aggravate cancer through creating immunosuppressive environments~\cite{desage_immune_2021} and stimulating cancer cell growth. The former occurs through secretion of immunosuppressive factors (e.g. TGF-$\beta$, IL-10, PGE2)~\cite{wong_modulating_2016, urso_detection_2021} and induction of regulatory T cells~\cite{hegmans_mesothelioma_2006}, the latter through secretion of growth factors (e.g. VEGF, FGF)~\cite{kumar-singh_angiogenic_1999} and influence on extracellular matrix (ECM) interactions~\cite{chu_immune_2019}.

On the other hand, experts noted that HGSC can increase mesothelioma abundance as it invades the abdominal cavity, where mesothelial cells naturally line the perotoneum. HGSC cells may induce their transformation into tumor-promoting phenotypes through mechanisms such as cell signaling pathways~\cite{islam_ovarian_2024, stadlmann_interactions_2014}, ECM interations~\cite{del_rio_ovarian_2021}, and influence on mesothelial-to-mesenchymal transition (MMT)~\cite{islam_ovarian_2024}. This bidirectional interplay makes possible a reinforcing cycle: mesothelioma cells promote HGSC growth, and HGSC invasion further increases mesothelioma abundance. Therefore, the observed trend does not establish that high mesothelioma presence is a definitive cause of HGSC, a crucial basis for considering their therapeutic potential. These complexities and possible alternative explanations underscore the need for more rigorous examination of the hypothesis.

\noindent \textbf{\modelname makes reasonable connections between different biological topics.} We observed that \modelname was able to progress from identifying individual statistical trends to performing high-level analysis on biological mechanisms, often through connecting different biological topics. On Dataset 4, \modelname focused on the significant increase of CD20 expression in tumor microenvironment-derived cytokines (TmDCc) ($\log_2 \text{FC} = 0.726746$, $\text{adjusted p-value} = 0.041301$), and proposed that this may indicate broader immune responses involving B and T Cell interactions, possibly mediated through CD20. 

Human evaluators stated how this hypothesis is supported by relevant research, but has not yet been directly studied. Particularly, CD20's role in B and T Cell interactions has been touched on by existing research. CD20 can influence B Cells' role as antigen-presenting cells (APCs)~\cite{kap_cd20_2014}, and its direct association with MHCII and CD40~\cite{pavlasova_regulation_2020} also suggest an impact on T Cell activation. CD20+ B Cells may influence T Cell differentiation through cytokine production (e.g. IL-2, IL-4) or recruit T Cells through chemokine release (e.g. CCL3, CCL4, CXCL10)~\cite{zhang_roles_2023}.
Additionally, CD20's crucial role in disease progression and therapeutics is reflected in research on anti-CD20 therapies, which can take effect through multiple mechanisms, including diminishing Regulatory T Cell populations~\cite{weber_bcell_2010}.
Considering that these mechanisms are yet to be studied under TmDC conditions, further exploring the proposed perspective may uncover novel mechanisms of interaction and deeper insights into disease responses. This result shows \modelname's familiarity of commonly considered research directions and its ability to extend its analysis beyond individual, surface trends.

\noindent \textbf{\modelname carries out logically coherent bioinformatic pipelines in clinical cohort analysis and generates reliable hypothesis.} While the above cases were all produced on SPDB datasets, we also noticed a large number of high-quality hypotheses among results from clinical cohort datasets. The data characteristics, common research themes, and analysis methods of clinical cohort data all differ considerably compared with SPDB datasets. Therefore, the following results indicate the versatility and flexibility of \modelname's system design.

On the GBM dataset (Dataset 11), \modelname performed a comprehensive multi-step investigation of protein expression levels on GBM tumor and normal tissues. It first executed differential expression analysis (DEA) between GBM and normal samples to identify SNX32, along with VIM, LIMA1, NES, and S100A6, as significantly differentially expressed proteins. These proteins were filtered using log fold changes ($|\log_2 \text{FC}| > 1$) and false discovery rates ($ \text{FDR} < 0.05$), then identified by the LLM as proteins of interest based on both logFC values and P values. \modelname centered the subsequent analysis on SNX32, using correlation analysis (CA) to reveal that both VIM and NES exhibit prominent negative correlations with SNX32 ($\text{correlation coefficient} < -0.3, p < 0.05$). It then conducted gene set enrichment analysis (GSEA), after which it highlighted the downregulation of synaptic vesticle docking and the upregulation of integrin-mediated signaling. By integrating these multiple layers of evidence, \modelname formulated the hypothesis that SNX32 functions as a tumor suppressor protein, primarily through its inhibitory effect on VIM expression.

According to human evaluators, the above process demonstrates that the system can not only understand and apply various bioinformatics tools, but also construct a logically coherent bioinformatics pipeline. Regarding the proposed hypothesis, SNX32 (Sorting Nexin 32) has never been studied in the context of GBM, as no relevant literature was found. In contrast, VIM (Vimentin) is a well-studied structural protein that plays a crucial role in tumor progression by promoting epithelial-mesenchymal transition~\cite{serrano-gomez_regulation_2016}. It has been reported to be a critical marker for glioma progression, associated with increased tumor invasiveness and poor prognosis ~\cite{nowicki_proteomic_2019, liu_vimentin_2023}. The GSEA results from \modelname aligned well with these known trends, and correlation analysis results provided a basis for connecting SNX32 and VIM. Although further experimental validation is needed to establish the proposed causal relationship, experts believed that obtaining this novel and biologically plausible hypothesis under fully automated experimental settings effectively demonstrates \modelname's ability in knowledge discovery.

\noindent \textbf{Better leveraging the biological knowledge of LLMs can potentially improve \modelname's performance.} In Dataset 1, cell conditions were labeled as "GvHD" or "Normal", without information on whether the disease is acute or chronic. As a result, \modelname drew broad conclusions regarding GvHD, reporting a significant decrease in the abundance of cytotoxic T cells and deducing that this trend may contribute to the impaired immune response in GvHD patients. However, this claim overlooked crucial biological distinctions ~\cite{jiang_t_2021, soares_naive_2019} between acute and chronic GvHD, regarding both Cytotoxic T Cell and Memory T Cell dynamics.

After pinpointing this flawed result, our subsequent experiments revealed that with accurate instructions, our base LLMs could correctly explain the above differences. This means that \modelname's current framework and prompt designs have not fully exploited the knowledge base of LLMs. In the above example, this limitation caused it to establish proposals on identified trends without fully considering pertinent biological context and nuances. We conclude that improving the rigor of the hypotheses would require \modelname to more effectively elicit knowledge from LLMs at key steps of its analysis process.

\section{Discussion}
In this paper, we introduced \modelname, an end-to-end, fully automatic system for scientific discovery from raw proteomics data. An LLM acts as the core coordinator of the system, performing hierarchical planning, analysis tool calling, iterative feedback and refinement, and hypothesis proposal. We incorporated a large number of professional bioinformatics tools and organized them into analysis workflows that can be conveniently called by the system to investigate specific datasets. In this way, we have built upon the capabilities of the base LLM to form a system that better adheres to the empirically grounded and exploratory nature of scientific research.

We performed detailed evaluation on \modelname's outputs both quantitatively and qualitatively. We constructed a set of 5 metrics and corresponding instructions, then used LLMs to perform large-scale automatic evaluation on a total of 191 hypotheses from two proteomics dataset types. Detailed reviews and scoring from 4 human experts corroborated the reliability and rigor of our automatic evaluation method. Experts also identified a number of novel hypotheses that point out promising directions for further research. Through examining these notable cases, we highlighted \modelname's ability to pinpoint underexplored biological topics, couple specific quantitative results with general domain knowledge, and establish connections between multiple statistical trends or biological entities. Capabilities such as these empower the system to progress past surface-level observations to perform in-depth scientific reasoning and discovery. In general, results demonstrate that \modelname consistently produces reliable results, is capable of forming novel and insightful hypotheses, and can be easily adapted to different data types. Our work is the first to achieve both proteomics research and result evaluation in an effective, automated, and end-to-end manner. \modelname distinguishes itself from bioinformatics assistants that focus on single analysis steps, and through simulating the full scientific inquiry process, makes important advances towards new paradigms of bioinformatics research.

We identify several current limitations of \modelname. First, as elaborated in Section~\ref{case-studies}, facing flawed or incomprehensive results, \modelname has yet to fully elicit the base language model's knowledge to provide necessary explanations and qualifications. Refining prompting instructions within the system and augmenting it with LLM self-reflection mechanisms are potential methods for addressing this shortcoming. Second, since we provide the LLM with direct access to all previous analysis records, the number of workflows or tools called is limited by the context length of the language model. In most of our experiments, \modelname called a maximum of 5 workflows on SPDB datasets and 6-8 workflows on clinical cohort datasets. With the goal of enabling \modelname to handle more complicated analysis processes, a potential improvement is to develop more sophisticated memory management methods to concisely and dynamically provide the system with the most relevant analysis records at each given stage. Such improvements will potentially increase the complexity of the system's analysis by large margins and amplify the advantages of its iterative refinement mechanism, leading to more in-depth results.

We believe that \modelname charts a promising path towards more efficient and comprehensive research in bioinformatics. Its features, including using research objectives to guide analysis, flexibly calling professional analysis tools, and iteratively adjusting the research process, are highly generalizable to diverse directions of scientific discovery. Therefore, the design principles of \modelname can be extended beyond proteomics to multi-omics analysis or even other realms of biomedical research. Moreover, future developments in both general large language models and specialized bioinformatics analysis methods will continually improve the quality of \modelname's analysis and results.

\section{Method}

\subsection{Analysis Workflows and Tools}
In this section, we provide an overview of the main analysis workflows and tools available to \modelname in our main experiments, explaining the role of language models in flexibly and correctly configuring the tools.

\subsubsection{Analyzing CyTOF Data}
Workflows in this section are tailored towards analyzing proteomics data obtained from CyTOF sequencing.

\noindent \textbf{FlowSOM Clustering and Cell Type Annotation}
This workflow clusters single cells based on protein expression, extracts highly expressed cell marker proteins of each cluster, then performs cell type annotation on the clusters. For clustering, we use the \textsf{CATALYST}~\cite{crowell2024catalyst} package to execute the FlowSOM~\cite{quintelier_analyzing_2021} algorithm, a self-organizing map-based method designed for flow or mass cytometry data. We set the inital cluster number to 30. We use the \textsf{scran}~\cite{lun2016scran} package to identify the top 10 cell markers of each cluster to prepare for automatic cell type labeling.

Previous research~\cite{hou_assessing_2024} has shown that GPT4 can generate cell type annotations given cell markers and the tissue type, achieving higher degrees of agreement with human annotations compared with conventional reference-based approaches. Therefore, we similarly use GPT-4o for labeling and designed the following prompt:

\begin{myinfobox}{Prompt for Cell Type Annotation}
\label{infobox:celltype-annotation}
\color{textcolor}

Identify the cell type of <tissue name> using the following markers, arranged from highest to lowest expression levels. This means you should consider the first several markers in the list to be more important. Provide your result as the most specific cell type that is possible to be determined.

Markers: <markers>

Provide your output in the following format:

\ \ Analysis: <brief analysis of the cell markers and their relationship to cell types> Cell Type: <cell type name>

Strictly adhere to this format and do not include any additional words or explanations.

\end{myinfobox}

Furthermore, we included an additional annotation refinement step to correct cell type name discrepancies that arose from using multiple LLM calls for the initial annotation process:

\begin{myinfobox}{Prompt for Cell Type Annotations Refinement}
\label{infobox:celltype-refinement}
\color{textcolor}

You are a bioinformatics researcher. You will be given a list of <cell type number> cell type annotations. The annotations were performed individually, so there may be cases where the same cell type has been assigned slightly different names. Your task is to refine the list of cell type annotations to ensure that the same cell type is assigned the same name. For instance, if two annotations are 'Memory CD8$^{+}$ T Cells' and 'CD8$^{+}$ memory T cell' respectively, you may choose to change them both to 'Memory CD8$^{+}$ T Cells'. Additionally, if you think some names are too specific, you can choose to make them more general so that they can be merged with other cell types in the list to facilitate future analysis. For instance, if there are many annotations of 'Memory CD8$^{+}$ T Cells' and only one 'Central Memory CD8$^{+}$ T Cells', you may change the latter to 'Memory CD8$^{+}$ T Cells'. Ensure that the names are concise and specific. Provide your output in the same format as the input (a list of cell type annotations separated by commas, where each term is a refined cell type name). Do not include any additional words or explanations.

Original annotations: <original cell type annotations>

\end{myinfobox}

After cell type labeling, clusters that were assigned the same cell types were merged, and the final cell types were stored in the SingleCellExperiment object as an additional set of cluster codes. We also organize the full set of cell types and their corresponding cell markers used into a natural language description, which is stored in the system's history as the execution result of this workflow. Figure \ref{fig:celltype_workflow} illustrates the full process of this workflow.

\noindent \textbf{Clustering and Annotation Refinement}
This workflow provides the option to perform further clustering and more fine-grained cell type labeling on an existing cluster. It performs dimension reduction on the clustered cells based on protein expression levels, then generates a plot where the cells are colored according to cell type. The LLM interprets the plot and outputs one cell type to refine. We subset the data object to only include cells of the selected cell type, then implement the same clustering and labeling workflow as for the initial clusters. Finally, clusters and cell type labels are updated with the refined annotations.

For example, a large T Cell cluster may be refined into three sub-clusters: "CD4$^{+}$ T Cells", "CD8$^{+}$ T Cells", "Regulatory T Cells".

\noindent \textbf{Protein Abundance Visualization}
This workflow intends to provide general, qualitative information on the proteomic landscape of the dataset and includes analysis on two types of plots. First, it generates a heatmap of all protein expressions over different samples, with auxiliary labels of sample conditions provided. Second, it visualizes expression levels of individual proteins. Based on the research objective, the LLM selects several proteins of interest, and a plot is generated for each protein, with samples colored according to sample conditions. The LLM interprets both these images and generates a textual description of notable trends. Figure \ref{fig:vis_workflow} shows two example plots generated by running this workflow on a CyTOF dataset.

\noindent \textbf{Differential Abundance Analysis of Cell Types}
The following workflows focus on identifying differentially expressed biological molecules. We use the edgeR~\cite{chen2024edger} algorithm in the \textsf{diffcyt}~\cite{weber2024diffcyt} package to perform differential analysis of cell type abundances over different sample characteristics.

The LLM begins by selecting a metadata field to focus the analysis on, based on the research objective and data description, containing a list of available fields and their example values. Subsequently, it is given the full list of existing values of this metadata field and selects several contrasts to analyze. This step allows both comparing one group of cells against another and comparing one group against all remaining cells. We use these chosen parameters to construct a design matrix for calling \textsf{diffcyt}. The resulting data is stored in a csv file and interpreted by the LLM.

\noindent \textbf{Differential Expression Analysis of Proteins Stratified by Cell Type}
This workflow uses the limma~\cite{ritchie2015limma} algorithm in the \textsf{diffcyt} package to calculate differential protein expression stratified by cell type. We similarly prompt the LLM to specify cell groupings for comparison. Here we include an additional step, where the LLM selects a subset of cell types to focus its analysis on, based on the research objective. After executing limma, we only keep data entries on the selected cell types and call the LLM to generate textual data interpretation.

\noindent \textbf{Differential Expression Analysis of Proteins Over All Cells}
For this workflow, we first merge all cells into a single cluster, then follow the steps in the previous workflow. This allows the system to get information on protein expression differences over the entire set of cells.

\renewcommand{\figurename}{Fig.}
\renewcommand{\thefigure}{9}
\begin{figure}
    \centering
    \includegraphics[width=0.99\linewidth]{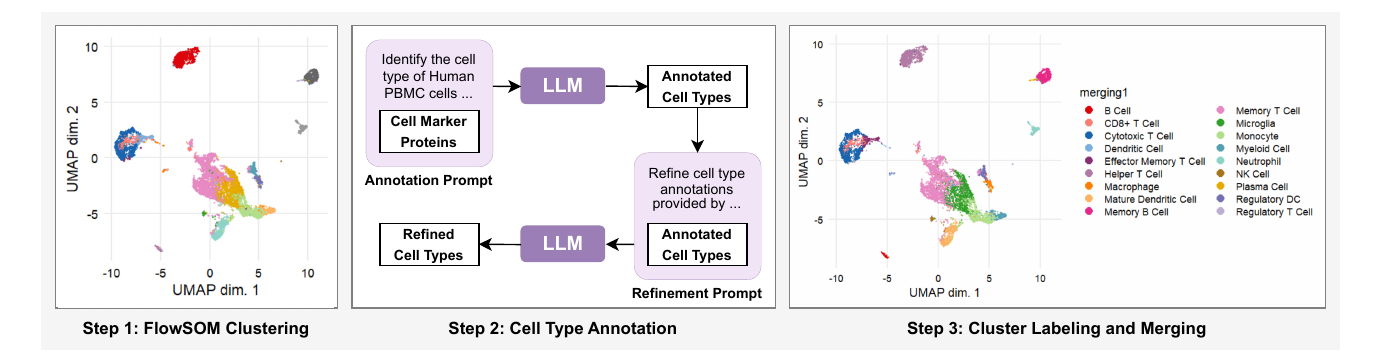}
    \caption{The steps in the workflow FlowSOM Clustering and Cell Type Annotation. The procedure consists of performing FlowSOM clustering based on protein expression data, annotating and refining cell types based on top cell markers in each cluster, and finally assigning cell types as cluster names and merging clusters with the same cell types. The dataset used was \textit{Immune phenotyping of diverse syngeneic murine brain tumors identifies immunologically distinct types}, downloaded from SPDB.}
    \label{fig:celltype_workflow}
\end{figure}

\renewcommand{\figurename}{Fig.}
\renewcommand{\thefigure}{10}
\begin{figure}
    \centering
    \includegraphics[width=0.99\linewidth]{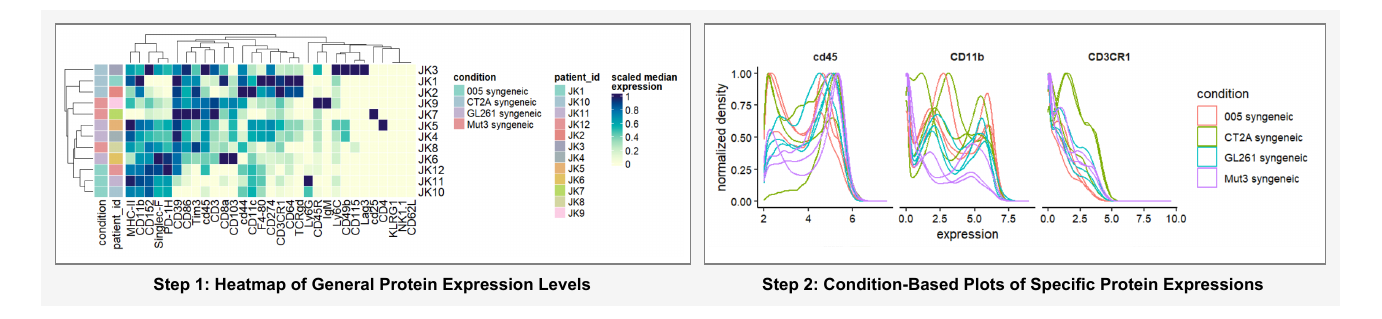}
    \caption{The steps in the workflow Protein Abundance Visualization. The workflow analyzes general protein abundance across different samples, then specific protein expression level distributions of a selected subset of proteins. Both plots are then interpreted by an LLM. The dataset used was \textit{Immune phenotyping of diverse syngeneic murine brain tumors identifies immunologically distinct types}, downloaded from SPDB.}
    \label{fig:vis_workflow}
\end{figure}

\subsubsection{Analyzing Clinical Cohort Data}
We provide a different set of workflows for \modelname to analyze clinical cohort data from mass spectrometry sequencing. All workflows excluding survival analysis and correlation analysis were conducted using the \textsf{BioEnricher}~\cite{liu2024bioenricher} package. Here we operate on a \textsf{BioEnricher} object that is constantly updated with each executed workflow, instead of a SingleCellExperiment object for CyTOF data. We create the data object using two csv files containing protein expression data and clinical metadata respectively, filtering out entries containing more than 25\% missing values and imputing all remaining missing values using kNN. The system automatically generates the data description from the raw csv files and takes no additional input information.

\noindent \textbf{Differential Expression Analysis}
For parameter selection of Differential Expression Analysis (DEA), we adopt a similar procedure as differential analysis for CyTOF data, allowing \modelname to select a single metadata field, followed by one or more sets of conditions for comparison. We use the limma algorithm in \textsf{BioEnricher} to calculate top up-regulated and down-regulated proteins, as well as relevant statistics such as log fold changes and P values.

\noindent \textbf{Consensus Clustering}
This workflow clusters samples into several subtypes based on their proteomic profiles. We use all proteins in the data and perform clustering using the non-negative matrix factorization (NMF) algorithm and Euclidean distances. \textsf{BioEnricher} clusters samples into 2 to 4 subtypes, then selects the optimal clustering result by aggregating multiple metrics of clustering quality, such as Calinski-Harabasz index, PAC, and Davies-Bouldin index. After executing the clustering algorithm, we add the final subtype indices to the data object as an additional metadata field.

\noindent \textbf{Enrichment Analysis}
The Enrichment Analysis workflow first runs DEA on the data object if it has not already been performed. Based on the differentially expressed proteins, we identify enriched biological pathways by performing both over-representation analysis (ORA) and gene-set enrichment analysis (GSEA). Available pathway databases include GO, KEGG, Wiki, Reactome, MsigDB, etc. The result files of both algorithms are jointly provided to the LLM for analysis.

\noindent \textbf{Survival Analysis}
We implement two types of survival analysis using the Python package \textsf{lifelines}~\cite{davidson-pilon_lifelines_2024}. First, we perform survival analysis on discrete classifications of low or high protein expression levels using log rank tests. We adjust the threshold for classification between the 20 and 80 percentiles of the protein expression data with intervals of 10 and select the threshold that yields the lowest p value for the final results. Second, we directly analyze continuous expression data using cox univariate regression and record key statistics such as correlation coefficients and p values.

When calling this workflow, \modelname selects the type of analysis to perform, as well as parameters including the metadata field to use for survival time, the field to use for event status, and a list of key molecules to focus on.

\noindent \textbf{Correlation Analysis on Clinical Features}
We implement correlation analysis in Python using \textsf{scipy} and mainly use Pearson correlation. This workflow calculates correlation levels between the expression of any molecule in the data and a clinical feature in the metadata. \modelname specifies the list of molecules, molecule type, and clinical trait to analyze when calling the workflow.

\noindent \textbf{Correlation Analysis Between Biological Molecules}
We similarly use \textsf{scipy} to investigate correlations between different biological molecules of same or different types (proteins, RNAs, phosphoproteins, etc.) This makes the workflow useful for both single and multi omics scenarios. \modelname calls the workflow by selecting two lists of molecules to analyze and their correspondingm molecule types.

\subsubsection{Accessing External Data}
We include additional workflows for \modelname to reference external information from datasets or databases based on biological molecules and diseases of interest.

\noindent \textbf{Correlation Analysis on External Datasets}
We use the \textsf{DataChat}~\cite{liu2024datachat} package for integrated usage of external datasets such as The Cancer Genome Atlas (TCGA, https://www.cancer.gov/ccg/research/genome-sequencing/tcga) and the Clinical Proteomic Tumor Analysis Consortium (CPTAC)~\cite{thangudu_abstract_2020}. Given the data description, \modelname searches for the required cancer type within available datasets in \textsf{DataChat}, locating the most relevant source database and datasets. \textsf{DataChat} provides diverse functions for correlation analysis between different molecule types (proteins, RNAs, phosphoproteins, etc.) \modelname selects relevant molecules and specifies molecule types to run correlation analysis, then automatically interprets the resulting plot.

\noindent \textbf{Survival Analysis on External Datasets}
This workflow is similarly based on an automatically selected dataset in \textsf{DataChat}. \modelname selects biological molecules on which to perform survival analysis and specifies the molecule type. The workflow then produces a Kaplan-Meier plot comparing the survival of patients with low and high expressions of the selected molecule, labeled with the calculated P value. The LLM judges whether the moelcule has significant impact on survival based on the plot. Both of these workflows intend to enhance the quality of hypotheses proposed by \modelname through providing an avenue to corroborate preliminary observations using other data sources. In most cases, we observe that \modelname correctly chooses data analysis workflows first, followed by external data validation workflows, and is able to identify notable, relevant molecules when calling \textsf{DataChat}.

\noindent \textbf{The Human Protein Atlas (THPA)}
Additionally, we design a workflow based on the THPA~\cite{uhlen_tissue-based_2015} API (proteinatlas.org) to provide \modelname with general biological information on a comprehensive set of human proteins. The workflow first calls the LLM to select a protein of interest based on its analysis history, then uses the API to fetch the following information: related protein classes; related biological pathways; related molecular functions; cancers in which the protein has favorable prognosis; cancers in which the protein has unfavorable prognosis.

\subsection{Prompt Engineering in Automatic Evaluation}
\label{method-evaluation}

Here we provide the full prompts we used for performing automatic scoring on each of the 5 metrics.

\begin{myinfobox}{Prompt for Auto-evaluation: Paper Alignment}
\label{infobox:paper-align}
\color{textcolor}

Conduct a thorough comparison between the AI-generated conclusion and the conclusion from an original research paper in the context of proteomics research. Assess the degree of concordance in terms of:

\ \ a) Identification of key protein markers or cell types

\ \ b) Reported biological processes or mechanisms

\ \ c) Statistical tests and quantitative results (e.g., fold changes, p-values)

\ \ d) Quality and alignment of statistical analysis workflows

\ \ e) Proposed final conclusion or hypothesis

\ \ f) Implications for the field of study

Score on a scale of 0-5:

\ \ 0: No alignment; AI conclusion contradicts or misses all key points from the original

\ \ 1: Minimal alignment; only superficial similarities in general topic

\ \ 2: Partial alignment; some key proteins, cell types, or biological conditions match, but significant discrepancies in main findings

\ \ 3: Moderate alignment; major findings and trends match, but differences in fine-grained names of protein markers or cell types, or divergences in deductions for further biological implications. Conclusions whose specific cell types overlap with those in the paper but exhibit notable differences should be in this category.

\ \ 4: Strong alignment; matches in main findings, key proteins or cell types, and most interpretations, with only minor differences in emphasis or detail. Conclusions whose specific cell types are close to those in the paper, with only minor differences, for instance in specificity, should be in this category.

\ \ 5: Perfect alignment; AI conclusion captures all main points, key proteins and cell types, quantitative results, and biological interpretations from the original paper

AI-generated conclusion:
[Insert AI conclusion here]

Original paper conclusion:
[Insert original conclusion here]

Provide your output in the following format:

List of matching and divergent points:

General assessment:

Score (0-5): <0/1/2/3/4/5>

\end{myinfobox}

\begin{myinfobox}{Prompt for Auto-evaluation: Literature-Based Alignment}
\label{infobox:literature-align}
\color{textcolor}

Perform a comprehensive literature review using PubMed to evaluate a provided AI-generated conclusion's alignment with existing proteomics research. Consider:

\ \ a) Consistency with established trends in protein or cell type abundances with respect to similar biological condition comparisons

\ \ b) Concordance with previously reported quantitative statistical results

\ \ c) Consistency with systems biology perspectives and further general implications in the field

Score on a scale of 0-5:

\ \ 0: Contradicts well-established proteomics findings; multiple studies refute the conclusion

\ \ 1: Limited support; mostly contradicts current literature with only minor points of agreement

\ \ 2: Mixed support; some aspects align with literature but significant contradictions exist

\ \ 3: Moderate support; generally aligns with literature, but some notable discrepancies or gaps. Conclusions whose specific cell types overlap with those in existing papers but exhibit notable differences should be considered to have notable gaps.

\ \ 4: Strong support; aligns well with multiple studies, only minor inconsistencies. Conclusions whose specific cell types are close to those in existing papers, with only minor differences, for instance in specificity, should be considered to have minor inconsistencies.

\ \ 5: Excellent support; perfectly aligns with well-established findings across multiple studies and reviews

PubMed Articles:
[Insert relevant PubMed article information here]

AI-generated conclusion:
[Insert AI conclusion here]

Provide your output in the following format:

Key supporting studies (with PMIDs):

Key contradicting studies (if any, with PMIDs):

Gaps in current literature relevant to the conclusion:

General assessment:

Score (0-5): <0/1/2/3/4/5>

\end{myinfobox}

\begin{myinfobox}{Prompt for Auto-evaluation: Literature-Based Novelty}
\label{infobox:literature-novelty}
\color{textcolor}

Conduct a thorough PubMed search to evaluate the novelty of the AI-generated conclusion in the context of proteomics research. Consider:

\ \ a) Identification of previously unknown disease biomarkers or immune signatures

\ \ b) Novel insights into protein functions, cell type functions, biological pathways or mechanisms

\ \ c) Unique integration of proteomics data analysis results with general proteomics and biological knowledge

\ \ d) Innovative approaches to data interpretation in proteomics

\ \ e) Potential for opening new avenues of research in the field

Score on a scale of 0-5:

\ \ 0: Entirely unoriginal; all aspects have been extensively reported in multiple studies

\ \ 1: Minimal novelty; mostly reiterates known findings with only trivial new aspects

\ \ 2: Modest novelty; combines known concepts in a somewhat new way, but no significant new insights

\ \ 3: Moderate novelty; presents a fresh perspective on well-studied proteomics concepts or ideas

\ \ 4: High novelty; uncovers a previously unreported trend, idea, or interpretation in proteomics research

\ \ 5: Groundbreaking; presents an entirely new concept or approach that could significantly advance the field

PubMed Articles:
[Insert relevant PubMed article information here]

AI-generated conclusion:
[Insert AI conclusion here]

Provide your output in the following format:

Most closely related existing research (with PMIDs):

Aspects that distinguish this conclusion from existing work:

Potential impact on future proteomics research:

General assessment:

Score (0-5): <0/1/2/3/4/5>

\end{myinfobox}

\begin{myinfobox}{Prompt for Auto-evaluation: Logical Coherence}
\label{infobox:logical-coherence}
\color{textcolor}

Assess the logical coherence and biological plausibility of a provided AI-generated proteomics conclusion based on fundamental principles of molecular biology, biochemistry, bioinformatics and proteomics. Evaluate:

\ \ a) Consistency with known protein or cell type functions

\ \ b) Adherence to established biological mechanisms and characteristics existent in the emphasized disease(s)

\ \ c) Plausibility of proposed molecular mechanisms

\ \ d) General logical coherence and consistency

Score on a scale of 0-5:

\ \ 0: Fundamentally flawed; violates basic principles of molecular biology or biochemistry

\ \ 1: Major logical inconsistencies; proposed mechanisms highly unlikely based on current biological knowledge

\ \ 2: Some logical gaps; parts of the conclusion are biologically plausible, but significant aspects are questionable

\ \ 3: Generally sound; mostly adheres to biological principles with a few minor logical leaps

\ \ 4: Logically robust; aligns well with biological principles, only very minor questionable points

\ \ 5: Exemplary logical coherence; fully adheres to all relevant biological principles and considers potential complexities in proteomics data interpretation

AI-generated conclusion:
[Insert AI conclusion here]

Provide your output in the following format:

Strengths in biological reasoning:

Weaknesses or questionable aspects:

Suggestions for improving biological plausibility:

General assessment:

Score (0-5): <0/1/2/3/4/5>

\end{myinfobox}

\begin{myinfobox}{Prompt for Auto-evaluation: Evaluability}
\label{infobox:evaluability}
\color{textcolor}

Assess the degree to which the AI-generated conclusion can be effectively evaluated based on current scientific knowledge, available data, and the nature of the claim. Consider the following factors:

\ \ a) Clarity and specificity of the conclusion

\ \ b) Adherence to statistical trends or biological conclusions presented in the original paper

\ \ c) Existence of established methods to test the claim

\ \ d) Presence of related studies in the literature

\ \ e) Technological feasibility of verifying the conclusion through either external data or further experimentation

\ \ f) Time frame required for potential validation (short-term vs. long-term implications)

\ \ g) Ethical considerations for testing the conclusion

Score on a scale of 0-5:

\ \ 0: Not evaluable; Conclusion is too vague, ambiguous, or poorly formulated to be evaluated. No relevant data or established methods exist to assess the claim. Contradicts fundamental scientific principles or ethical considerations.

\ \ 1: Minimally evaluable; Conclusion is mostly unclear but contains some assessable elements. Very limited relevant data or literature available. Would require significant technological advancements to test.

\ \ 2: Partially evaluable; Conclusion is somewhat clear but lacks crucial details. Some relevant data and methods exist, but significant gaps remain. Current methods can partially assess the claim, but with major limitations.

\ \ 3: Moderately evaluable; Conclusion is mostly clear and specific. Mostly sufficient relevant data and methods are available for a satisfactory evaluation. Current methods can largely assess the claim, with some limitations.

\ \ 4: Highly evaluable; Conclusion is clear, specific, and well-formulated. Sufficient relevant data and methods are available for comprehensive evaluation. Established methods can thoroughly assess most aspects of the claim.

\ \ 5: Fully evaluable; Conclusion is exceptionally clear, specific, and comprehensive. Extensive relevant data, literature, and established methods are readily available for well-rounded, in depth evaluation. Claim can be fully assessed with current scientific knowledge and technology.

AI-generated conclusion:
[Insert AI conclusion here]

Provide your output in the following format:

Key factors influencing evaluability:

Suggested evaluation procedure, including necessary data and experimentation methods:

Challenges in evaluation (if any):

Suggestions for improving evaluability:

General Assessment:

Score (0-5): <0/1/2/3/4/5>

\end{myinfobox}

\backmatter

\newpage
\bibliography{sn-bibliography}

\clearpage
\begin{appendices}

\section{List of Datasets}
\label{list-of-datasets}
Table~\ref{tab:dataset-info} lists information about the 12 datasets we used for our main experiments. The first 10 datasets are CyTOF datasets directly downloaded from SPDB. Datasets 11 and 12 are MS datasets on clinical cohorts of GBM and HCC, respectively.

\input{table-dataset-info}

\section{Human Evaluation Instructions}

We provided the following instructions to guide human experts during their evaluation. Experts were not required to complete each metric, but directly gave open-ended reviews on aspects of the hypotheses that they found notable.
\\

\rowcolors{2}{gray!10}{gray!10}
\centerline{
\begin{tabular}{m{10cm} 
                >{\centering\arraybackslash}m{2cm} 
                >{\centering\arraybackslash}m{2cm}}
    \toprule
    \textbf{Section A: General Evaluation} &  unhelpful &   helpful \\ 
    \midrule
    How would you rate the overall clarity and comprehensibility of the results presented? &  $\bigcirc$ &  $\bigcirc$ \\ \addlinespace[0.5em]
    Does \modelname seem to provide valuable insights into the research questions? &  $\bigcirc$ &  $\bigcirc$ \\ \addlinespace[0.5em]
    Do you think the use of \modelname is a promising approach for biological research? &  $\bigcirc$ &  $\bigcirc$ \\ 
    
    \bottomrule
\end{tabular}}
\centerline{
\begin{tabular}{m{10cm} 
                >{\centering\arraybackslash}m{2cm} 
                >{\centering\arraybackslash}m{2cm}}
    \toprule
    \textbf{Section B: Comparison with Published Papers} &  unhelpful &   helpful \\ 
    \midrule
    How well do the results from \modelname align with the findings in the corresponding published papers?  &  $\bigcirc$ &  $\bigcirc$ \\ \addlinespace[0.5em]
    Are there any significant differences between the automated analysis results and the scientist-obtained results from published papers? If so, please describe. &  $\bigcirc$ &  $\bigcirc$ \\ \addlinespace[0.5em]
    In case of any inconsistencies, based on your professional background, please determine whether it is trivial, incorrect, or if it represents a new important and reasonable research direction. If a valuable new research direction emerges, please elaborate on its novelty and importance. &  $\bigcirc$ &  $\bigcirc$ \\ \addlinespace[0.5em]
    Does the objective presented in the results comprehensively address the important research questions related to the corresponding published papers? Are there any additional research questions that you think should be addressed based on the results? &  $\bigcirc$ &  $\bigcirc$ \\ 
    
    \bottomrule
\end{tabular}}

\centerline{
\begin{tabular}{m{10cm} 
                >{\centering\arraybackslash}m{2cm} 
                >{\centering\arraybackslash}m{2cm}}
    \toprule
    \textbf{Section C: Objectives \& Hypotheses} &  unhelpful &   helpful \\ 
    \midrule
    Is the proposed objective reasonable? &  $\bigcirc$ &  $\bigcirc$ \\ \addlinespace[0.5em]
    Does the analysis in the hypotheses fulfill the planning of the objective? Is it evidence-based?  &  $\bigcirc$ &  $\bigcirc$ \\ \addlinespace[0.5em]
    Does the hypothesis present cell types and proteins that were not emphasized but are of great significance in the original paper? &  $\bigcirc$ &  $\bigcirc$ \\ \addlinespace[0.5em]
    How strongly logical is the automated analysis process presented in the entire results, from objective design to hypothesis formulation?&  $\bigcirc$ &  $\bigcirc$ \\ \addlinespace[0.5em]
    Calculate relevant indicators according to the original literature and original data to determine whether the cell type and marker are correctly corresponding, and confirm that the values of logFC and P-value in the key statistics are within a reasonable range and show the correct trend.&  $\bigcirc$ &  $\bigcirc$ \\ 
    
    \bottomrule
\end{tabular}}

\centerline{
\begin{tabular}{m{10cm} 
                >{\centering\arraybackslash}m{2cm} 
                >{\centering\arraybackslash}m{2cm}}
    \toprule
    \textbf{Section D: Hypotheses Generated} &  unhelpful &   helpful \\ 
    \midrule
    Are the hypotheses generated by \modelname reasonable and testable? &  $\bigcirc$ &  $\bigcirc$ \\ \addlinespace[0.5em]
    How scientific is the proposed biological hypothesis? &  $\bigcirc$ &  $\bigcirc$ \\ \addlinespace[0.5em]
    Do the hypotheses provide new directions for further research? &  $\bigcirc$ &  $\bigcirc$ \\ 
    
    \bottomrule
\end{tabular}}

\centerline{
\begin{tabular}{m{10cm} 
                >{\centering\arraybackslash}m{2cm} 
                >{\centering\arraybackslash}m{2cm}}
    \toprule
    \textbf{Section E: Suggestions for Improvement} &  unhelpful &   helpful \\ 
    \midrule
    What improvements or modifications would you suggest for \modelname to enhance its performance? Are there any additional features or capabilities that you think should be added to the system? &  $\bigcirc$ &  $\bigcirc$ \\ 
    
    \bottomrule
\end{tabular}}

\end{appendices}

\end{document}

%% file: table-workflow-info.tex
\begin{table}[ht]
    \caption{Overview of analysis workflows.}
    \label{tab:workflow_info}
    \begin{tabular}{p{0.08\textwidth}p{0.17\textwidth}|p{0.10\textwidth}p{0.1\textwidth}p{0.1\textwidth}|p{0.3\textwidth}}
        \toprule
        \textbf{Type} & \textbf{Workflow} & \textbf{Packages} & \textbf{Parameters} & \textbf{Output} & \textbf{Description} \\
        \midrule
        \multirow{6}{0.08\textwidth}{\vfill \textbf{Single-Cell Data}} 
            & Clustering and Annotation &  \textsf{CATALYST}, \textsf{scran} & None & SCE Object & Performs FlowSOM clustering, top protein marker identification, and cell type labeling for each cluster. \\ 
            & Annotation Refinement &  \textsf{CATALYST}, \textsf{scran} & Cell Type for Refinement & SCE Object & Further clusters the selected cell types and annotates the subclusters with refined cell names.\\
            & Visualization & \textsf{ggplot} & List of Proteins & Plots & Visualizes protein abundances for general analysis. \\
            & Differential Abundance (Cell Types) & \textsf{diffcyt} & Metadata Field, Contrasts & Files & Performs differential abundance analysis on all labeled cell types using the \textsf{edgeR} algorithm. \\
            & Stratified Differential Expression (Proteins) & \textsf{diffcyt} & Metadata Field, Contrasts & Files & Performs differential expression analysis on all proteins, stratified by cell type, using the \textsf{limma} algorithm. \\
            & Differential Expression (Proteins) & \textsf{diffcyt} & Metadata Field, Contrasts & Files & Performs differential expression analysis on all proteins, over all cells, using the \textsf{limma} algorithm. \\
            \midrule
        \multirow{6}{0.08\textwidth}{\vfill \textbf{Bulk Clinical Cohort Data}} 
            & Differential Expression Analysis & \textsf{BioEnricher} & Metadata Field, Contrasts & Files & Performs differential expression analysis to locate up or down regulated proteins, using the \textsf{limma} algorithm. \\
            & Consensus Clustering & \textsf{BioEnricher} & None & \textsf{BioEnricher} Object & Performs NMF clustering on the data samples using all protein expressions. Selects the ideal cluster number by ensembling numerous metrics. \\
            & Enrichment Analysis & \textsf{BioEnricher} & Metadata Field, Contrasts & Files & Performs enrichment analysis (over-representation analysis and gene-set enrichment analysis) based on differential expression analysis results. \\
            & Survival Analysis & \textsf{lifelines} & Analysis Type, Survival Time Field, Event Status Field & Files & Performs either discrete or continuous survival analysis to explore the relationship between proteins and patient survival. \\
            & Clinical Correlation & \textsf{scipy} & List of Molecules, Clinical Feature Field & Files & Computes the Pearson correlations between biological molecule expression and a selected clinical feature. \\
            & Molecule Correlation & \textsf{scipy} & Lists of Molecules & Files & Computes the pairwise Pearson correlations between two lists of biological molecules (proteins, RNAs, etc.) \\
        \midrule
        \multirow{3}{0.08\textwidth}{\vfill \textbf{External Datasets}}
        & Correlation Analysis & \textsf{DataChat} & Dataset Name, Lists of Molecules & Files & Analyzes the pairwise correlations of the selected biological molecules using the specified external dataset, often under the condition of a certain cancer type. \\
        & Survival Analysis & \textsf{DataChat} & Dataset Name, List of Molecules & Files & Performs survival analysis regarding the selected biological molecules using the external dataset. \\
        & THPA & None & Name of Protein & Text & Searches for basic biological information on a selected protein using the API of The Human Protein Atlas (THPA).\\
        \bottomrule
    \end{tabular}
\end{table}

%% file: table-llm-auto.tex
\begin{table}[ht]
    \centering

    \caption{Performance metrics of different large language models across various benchmarks.}
    \label{tab:general_domain_results}
    \begin{tabular}{p{1.8cm}p{4cm}|>{\centering\arraybackslash}p{1.8cm}>{\centering\arraybackslash}p{1.8cm}>{\centering\arraybackslash}p{1.8cm}>{\centering\arraybackslash}p{1.8cm}>{\centering\arraybackslash}p{1.8cm}}
        \toprule
        \multirow{2}{*}{\textbf{Domain}} & \multirow{2}{*}{\textbf{Models}} & \centering \textbf{MultiMedQA} & \multicolumn{2}{c}{\textbf{AlpacaEval 2}} & \textbf{MMLU} & \textbf{GPQA} \\
         & &  \textbf{Avg. Acc (\%)} & \textbf{LC (\%)} & \textbf{WR (\%)} & \textbf{Acc (\%)} & \textbf{Acc (\%)} \\
        \midrule
        \multirow{7}{*}{\textbf{General}} 
        & Mixtral-8x7B-Instruct~\cite{jiang2024mixtral}  &  63.17  &  23.70  &  18.30  &  70.60  & 39.50  \\
        & Mixtral-8x22B-Instruct~\cite{jiang2024mixtral} &  79.16  &  30.90  &  22.20  &  77.80  &  - \\
        & Qwen-1.5-72B-Chat~\cite{bai2023qwen}      &  70.24  &  36.60  &  26.50  &  75.60  &  39.40 \\
        & Qwen-2-72B-Chat~\cite{yang2024qwen2}        &  81.81  &  38.10  &  39.10  &  82.30  &  42.40 \\
        & DeepSeek-v2-Chat~\cite{liu2024deepseek}       &  77.90  &  -  &  -  &  78.50  &  - \\
        & Llama-3-70B-Instruct~\cite{dubey2024llama}   &  82.66  &  34.40  &  33.20  &  82.00  & 39.50  \\
        & Llama-3.1-70B-Instruct~\cite{dubey2024llama} &  84.05  &  38.10  &  39.10  &  86.00  &  46.70 \\
        & GPT-3.5-Turbo          &  67.70  &  19.30  &  9.20  &  70.00  &  28.10 \\
        & GPT-4-Turbo            &  87.00  &  50.00  &  50.00 &  86.40  &  49.10 \\
        \midrule
        \multirow{5}{*}{\textbf{BioMed}} 
        & Med42-70B~\cite{christophe2024med42}              &  70.74  &  -  &  -  &  -  &  - \\
        & OpenBioLM-70B~\cite{OpenBioLLMs}          &  86.06  &  30.80  &  31.00  &  60.10 & 29.20 \\
 
        & Med-PaLM 2 (ER)~\cite{singhal2023towards}        &  85.46  &  -  &  -  &  -  &  - \\
        & Llama-3-70B-UltraMed~\cite{zhang2024ultramedical}  &  85.84  &  33.00  &  32.10  &  77.20  &  39.70 \\ \midrule

        & \textbf{Llama-3.1-70B-UltraMed (Ours)}
                                 &  86.25  &  43.45  &  46.09  & 85.58  &  45.76 \\

        \bottomrule
    \end{tabular}
\end{table}

%% file: table-llm-medqa.tex
\begin{table}[t]
  \caption{Main results on medical multiple-choice questions (MultiMedQA)}
  \label{tab:ultramedical_main_results}
  \begin{tabularx}{\textwidth}{l|CCCCCCCCCC}
    \hline\addlinespace[2pt]
      \multirow{4}{*}{\textbf{Model \& Task}} & 
      \multirow{4}{*}{\begin{tabular}[c]{@{}c@{}}\textbf{MedQA}\\ \textbf{(US 4-opt)}\end{tabular}} &
      \multirow{4}{*}{\begin{tabular}[c]{@{}c@{}}\textbf{MedMCQA}\\ \textbf{(Dev)}\end{tabular}}  &
      \multirow{4}{*}{\begin{tabular}[c]{@{}c@{}}\textbf{PMQA}\\ \textbf{(Reason)}\end{tabular}} &
      \multicolumn{6}{c}{\textbf{MMLU}} &
      \multirow{4}{*}{\textbf{Avg.}}\\
    \cmidrule{5-10}
    & & & &
      \begin{tabular}[c]{@{}c@{}}\textbf{Clinical}\\ \textbf{knowledge}\end{tabular} &
      \begin{tabular}[c]{@{}c@{}}\textbf{Medical}\\ \textbf{genetics}\end{tabular} &
      \textbf{Anatomy} &
      \begin{tabular}[c]{@{}c@{}}\textbf{Profess.}\\ \textbf{medicine}\end{tabular} &
      \begin{tabular}[c]{@{}c@{}}\textbf{College}\\ \textbf{biology}\end{tabular} &
      \begin{tabular}[c]{@{}c@{}}\textbf{College}\\ \textbf{medicine}\end{tabular} & \\
    \midrule
    Mixtral-8x7B-Instruct 
        &  52.8  & 49.7  &  46.2  &  71.7  &  70.0   &  62.2  &  71.0   &  77.8 &  67.1 & 63.17 \\
    Mixtral-8x22B-Instruct  
        &  73.1  & 63.3  &  71.4  &  84.2  &  89.0   &  77.0  &  88.2   &  88.2 &  78.0 & 79.16 \\
    Qwen1.5-72B-Chat 
        &  63.6  & 59.0  &  32.4  &  78.9  &  80.0   &  68.9  &  82.7   &  91.0 &  75.7 & 70.24 \\
    Qwen2-72B-Chat 
        &  75.3  & 66.6  &  68.8  &  85.7  &  93.0   &  80.7  &  89.7   &  94.4 &   82.1 & 81.81 \\
    DeepSeek-v2-Chat 
        &  68.6  & 61.5  &  71.0  &  83.0  &  90.0   &  73.3  &  86.8  &  88.9 &  78.0 & 77.90 \\
    Llama-2-70B-Chat
        &  47.3  & 41.9  &  63.8  &  64.9  &  70.0 &  54.1  &  59.2  &  66.7 &  61.3 & 58.80 \\
    Llama-3-70B-Instruct
        &  79.9 &  69.6  &  75.8  &  87.2  &  93.0 &  76.3  &  88.2  &  92.4 &  81.5 & 82.66 \\
    Llama-3.1-70B-Instruct
        &  81.4 &  72.2  &  76.8  &  85.1  &  95.3 &  80.4  &  91.4  &  93.1 &  80.8  & 84.05 \\
    GPT-3.5-Trubo 
        &  57.7  &  72.7  & 53.8 & 74.7 & 74.0 & 65.9 & 72.8 & 72.9  &  64.7 & 67.70 \\
    GPT-4-base (5-shot) 
        &  86.1  &  73.7 &  80.4 &  88.7 &  97.0 & 85.2 & 93.8 &  97.2  &  80.9 &  87.00 \\
    \midrule
    Med42-70B &  66.6  & 60.6  &  67.2  &  76.6  &  77.0   &  66.7  &  79.8   &  75.7 &  66.5 & 70.74 \\
    OpenBioLLM-70B  & 78.2  &  74.0  &  79.0  & 92.9  & 93.2  & 83.9   &  93.8 & 93.8  &  85.7   &  86.06 \\
    Med-PaLM 2 (ER)        &  85.4  &  72.3  & 75.0 &88.7 & 92.0 & 84.4 & 92.3 & 95.8 &83.2 & 85.46 \\
    Llama-3-70B-UltraMed & 84.8 & 73.2 & 80.0 & 86.8 & 92.0 & 84.4 & 93.8 & 93.1 & 84.4 & 85.84 \\
    \midrule
    \textbf{Llama-3.1-70B-UltraMed} & 85.2 &  72.9  &  77.8  &  87.2   &   95.0  &   83.7   &   94.9    &  95.1   &  84.4   & 86.25 \\
    \bottomrule
    \end{tabularx}
    \label{tab:med}
\end{table}

%% file: table-stats.tex
\renewcommand{\figurename}{Table}
\renewcommand{\thefigure}{4}

\begin{figure}[ht]
    \centering
    \begin{subfigure}[t]{0.45\textwidth}
        \centering
        \caption{Statistics of the datasets, workflows, and hypotheses used in our evaluation.}
        \begin{tabular}{lc}
            \toprule
            \# Datasets & 12 \\ 
            \quad \# Cytometry by Time-of-flight & 10 \\ 
            \quad \# Clinical Cohort & 2 \\ 
            \midrule
            \# Workflows & 15 \\ 
            \midrule
            \# Hypotheses & 191 \\ 
            \quad \# Cytometry by Time-of-flight & 147 \\ 
            \quad \# Mass Spectrometry & 44 \\ 
            \bottomrule
        \end{tabular}
        \label{tab:stats}
    \end{subfigure}
    \hfill
    \begin{subfigure}[t]{0.45\textwidth}
        \centering
        \caption{Frequency of workflow execution over one experimental run on 10 CyTOF datasets.}
        \begin{tabular}{cc}
            \toprule
            \textbf{Workflow} & \textbf{Frequency}\\ 
            \midrule
            Clustering and Annotation & 46\\
            Annotation Refinement & 21\\
            Visualization & 7\\
            Differential Abundance & 23\\
            Stratified Differential Expression & 28\\
            Differential Expression & 9\\
            THPA & 10\\
            \bottomrule
        \end{tabular}
        \label{tab:workflow-frequency}
    \end{subfigure}
    \caption{Summary statistics and workflow execution frequency}
    \vspace{-0.4cm}
    \label{fig:combined}
\end{figure}

%% file: table-average-scores.tex
\renewcommand{\tablename}{Table}
\renewcommand{\thetable}{6}

\begin{wraptable}{r}{0.6\textwidth}
    \centering
    \normalsize
     \vspace{-0.0cm}
      \caption{Overview of the automatic evaluation on 5 metrics.}
    
    \begin{tabular}{lccccc}
    \toprule
         \textbf{Metric} & \textbf{Mean} & \textbf{Median} & \textbf{Std} & \textbf{Max} & \textbf{Min}\\ \midrule
         Paper-based Alignment & 2.13 & 2 & 0.85 & 4 & 0 \\
         Literature-based Alignment & 3.24 & 4 & 1.12 & 5 & 0 \\
         Literature-based  Novelty & 3.29 & 3 & 0.52 & 4 & 0 \\
         Logical Coherence & 3.32 & 3 & 0.59 & 4 & 1 \\
         Evaluability & 3.55 & 4 & 0.50 & 4 & 3 \\
         \bottomrule
    \end{tabular}
    \vspace{-0.3cm}
    \label{tab:overview}
\end{wraptable}

%% file: table-dataset-scores.tex
\renewcommand{\tablename}{Table}
\renewcommand{\thetable}{5}
\begin{table}[]
\centering
\normalsize
\caption{Average scores on 5 metrics across all datasets. Detailed dataset information is provided in Appendix~\ref{list-of-datasets}}
\begin{tabular}{lcccccc}
\toprule
    Dataset & \# Results & Paper & Literature & Novelty &  Coherence & Evaluability \\ \midrule
    \multicolumn{7}{c}{ \textit{SPDB (CyTOF)}}
\\ \midrule
 
    Human PBMCs, Leukemia \cite{hartmann2019comprehensive} & 13 & 2.46 & 3.70 & 3.31 & 3.46 & 3.62\\
    Human HGSC tumor cells \cite{gonzalez2018commonly} & 15 & 1.80 & 2.73 & 2.87 & 3.20 & 3.57\\
    Human PBMCs, ICC \cite{wu2021distinct} & 15 & 2.00 & 2.60 & 3.40 & 3.00 & 3.07\\
    Human T cells \cite{suwandi2020multidimensional} & 15 & 2.27 & 3.33 & 3.13 & 3.07 & 3.40\\
    Human HCC tumor cells \cite{zheng2020trajectory}  & 15 & 2.13 & 3.40 & 3.40 & 3.27 & 3.60\\
    Mouse GBM tumor cells \cite{khalsa2020immune} & 15 & 2.67 & 3.40 & 3.27 & 3.53 & 3.67\\
    Human PBMCs, DLBCL \cite{shi2022single-cell} & 15 & 1.93 & 2.93 & 3.33 & 3.27 & 3.53\\
    Human GC cells \cite{wei2022immune} & 14 & 1.86 & 3.14 & 3.36 & 3.64 & 3.86\\
    Human HIV cells \cite{ma2022single-cell} & 15 & 2.20 & 3.33 & 3.33 & 3.47 & 3.33\\
    Human PBMCs, thyroid disease \cite{stensland2022peripheral} & 15 & 1.80 & 2.73 & 3.33 & 3.27 & 3.67\\ 
    \midrule
    \multicolumn{7}{c}{ \textit{Clinical Cohorts (MS)}}\\
    \midrule
    Human GBM tissues \cite{oh2020integrated} & 24 & 2.00 & 3.25 & 3.29 & 3.17 & 3.54\\
    Human HCC tissues \cite{jiang2019proteomics} & 20 & 2.40 & 4.00 & 3.40 & 3.60 & 3.70\\
    \bottomrule
\end{tabular}
\label{table:data_scores}
\end{table}

%% file: table-dataset-info.tex
\begin{table}[ht]
    \centering
    \caption{Information on the 12 proteomics datasets used.}
    \label{tab:dataset-info}
    \begin{tabular}{p{0.02\textwidth}|p{0.45\textwidth}|p{0.10\textwidth}|p{0.13\textwidth}|p{0.13\textwidth}}
        \toprule
        \textbf{\#} & \textbf{Dataset Name} & \textbf{Species} & \textbf{Tissue} & \textbf{Disease} \\
        \midrule
        1 & Comprehensive Immune Monitoring of Clinical Trials to Advance Human Immunotherapy\cite{hartmann2019comprehensive} & Homo Sapiens & Blood & Leukemia \\
        2 & Commonly Occurring Cell Subsets in High-Grade Serous Ovarian Tumors Identified by Single-Cell Mass Cytometry\cite{gonzalez2018commonly} & Homo Sapiens & Tumor & High-Grade Serous Ovarian Cancer \\
        3 & Distinct Immune Signatures in Peripheral Blood Predict Chemosensitivity in Intrahepatic Cholangiocarcinoma Patients\cite{wu2021distinct} & Homo Sapiens & Blood & Intrahepatic Cholangiocarcinoma \\
        4 & Multidimensional Analyses of Proinsulin Peptide-Specific Regulatory T Cells Induced by Tolerogenic Dendritic Cells\cite{suwandi2020multidimensional} & Homo Sapiens & T Cells & None \\
        5 & Trajectory and Functional Analysis of PD-1high CD4+CD8+ T Cells in Hepatocellular Carcinoma by Single-Cell Cytometry and Transcriptome Sequencing\cite{zheng2020trajectory} & Homo Sapiens & Tumor & Hepatocellular Carcinoma \\
        6 & Immune Phenotyping of Diverse Syngeneic Murine Brain Tumors Identifies Immunologically Distinct Types\cite{khalsa2020immune} & Mus Musculus & Tumor & Glioblastoma \\
        7 & Single-Cell Phenotypic Profiling to Identify a Set of Immune Cell Protein Biomarkers for Relapsed and Refractory Diffuse Large B Cell Lymphoma: A Single-Center Study\cite{shi2022single-cell} & Homo Sapiens & Blood & Diffuse Large B-Cell Lymphoma \\
        8 & Immune Profiling in Gastric Cancer Reveals the Dynamic Landscape of Immune Signature Underlying Tumor Progression\cite{wei2022immune} & Homo Sapiens & blood / tumor / non-tumor adjacent tissues & Gastric Cancer \\
        9 & Single-Cell Glycomics Analysis by CyTOF-Lec Reveals Glycan Features Defining Cells Differentially Susceptible to HIV\cite{ma2022single-cell} & Homo Sapiens & blood / endometrium / tumor tissues & HIV \\
        10 & Peripheral Immunophenotyping of AITD Subjects Reveals Alterations in Immune Cells in Pediatric vs Adult-Onset AITD\cite{stensland2022peripheral} & Homo Sapiens & Blood & Autoimmune Thyroid Disease \\
        \midrule
        11 & Integrated Pharmaco-Proteogenomics Defines Two
Subgroups in Isocitrate Dehydrogenase Wild-Type
Glioblastoma with Prognostic and Therapeutic
opportunities~\cite{oh2020integrated} & Homo Sapiens & tumor / non-tumor adjacent tissues & GBM \\
        12 & Proteomics Identifies New Therapeutic Targets of
Early-Stage Hepatocellular Carcinoma~\cite{jiang2019proteomics} & Homo Sapiens & tumor / non-tumor adjacent tissues & HCC \\
        \bottomrule
    \end{tabular}
\end{table}